\documentclass[10pt,twocolumn,letterpaper]{article}

\usepackage{cvpr}              

\usepackage[dvipsnames]{xcolor}

\usepackage{amsmath}
\usepackage{multirow}

\usepackage{amssymb}
\usepackage{makecell}
\usepackage{xcolor}
\newtheorem{proposition}{Proposition}
\newtheorem{observation}{Observation}
\newtheorem{definition}{Definition}
\definecolor{cvprblue}{rgb}{0.21,0.49,0.74}
\usepackage[pagebackref,breaklinks,colorlinks,citecolor=cvprblue]{hyperref}

\def\paperID{10881} 
\def\confName{CVPR}
\def\confYear{2024}

\title{Structured Gradient-based Interpretations via \\Norm-Regularized Adversarial Training}

\author{Shizhan Gong, Qi Dou, Farzan Farnia\\
The Chinese University of Hong Kong\\
{\tt\small \{szgong22, qdou, farnia\}@cse.cuhk.edu.hk}}

\begin{document}
\maketitle
\begin{abstract}
Gradient-based saliency maps have been widely used to explain the decisions of deep neural network classifiers. However, standard gradient-based interpretation maps, including the simple gradient and integrated gradient algorithms, often lack desired structures such as sparsity and connectedness in their application to real-world computer vision models. A frequently used approach to inducing sparsity structures into gradient-based saliency maps is to alter the simple gradient scheme using sparsification or norm-based regularization. A drawback with such post-processing methods is their frequently-observed significant loss in fidelity to the original simple gradient map. In this work, we propose to apply adversarial training as an in-processing scheme to train neural networks with structured simple gradient maps. We show a duality relation between the regularized norms of the adversarial perturbations and gradient-based maps, based on which we design adversarial training loss functions promoting sparsity and group-sparsity properties in simple gradient maps. We present several numerical results to show the influence of our proposed norm-based adversarial training methods on the standard gradient-based maps of standard neural network architectures on benchmark image datasets\footnote{The paper's code is available at: \url{https://github.com/peterant330/AdvGrad}}.

\end{abstract}    
\section{Introduction}

Deep neural networks have attained remarkable results in various computer vision tasks including image classification~\cite{krizhevsky2012imagenet}, object detection~\cite{zhao2019object}, and semantic segmentation~\cite{guo2018review}. However, understanding and explaining the decision-making process of these complex models remains a challenge, which is required for high-risk applications such as medical imaging~\cite{shen2017deep},  autonomous driving~\cite{kim2017interpretable}, and face recognition~\cite{wang2021deep}. To interpret the predictions of neural network classifiers, several explanation methodologies have been proposed in the literature. Among these explanation methods, \emph{gradient-based saliency maps} have been frequently applied to interpret image classifiers. The saliency maps provide insight into the regions of the input image contributing to the classifier's decision and potentially prompt the empirical understanding of phenomena from data-driven models.

\begin{figure}[tp!]
    \centering
    \includegraphics[width=\linewidth]{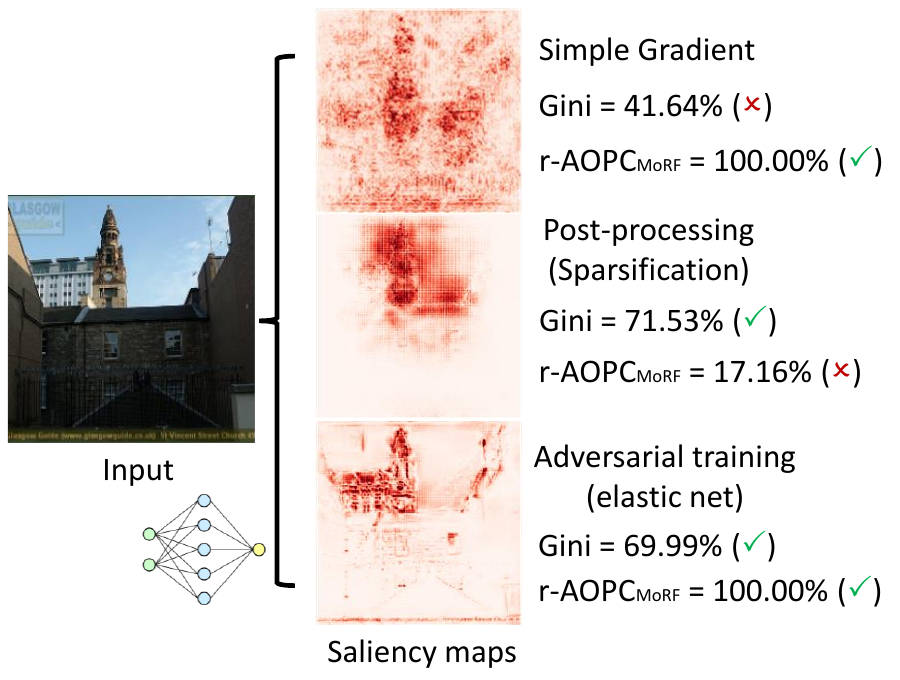}
    \caption{The original simple gradient map could be dense and noisy. Post-processing methods such as sparsification enhance the sparsity at the expense of lower fidelity to the original map. Our proposed in-processing strategy with adversarial training results in higher sparsity without losing fidelity to the simple-grad map. (We use the Gini index as a sparsity measure, and relative $\text{AOPC}_\text{MoRF}$ with respect to simple gradient as a fidelity score.)}
    \label{fig:teaser}
    \vspace{-4mm}
\end{figure}

Despite the widespread use of standard gradient-based interpretation maps such as simple gradients~\cite{simonyan2013deep} and integrated gradients~\cite{sundararajan2017axiomatic}, they often lack desired structures such as sparsity and connectedness. When applied to real-world computer vision tasks, the generated saliency maps have been frequently observed to be noisy and dense. Such unstructured behavior could hinder the application of gradient-based maps for detecting the influential features in the classifier's decision-making process. To obtain more structured saliency maps, several post-processing schemes~\cite{ smilkov2017smoothgrad,levine2019certifiably,zhang2023moreaugrad,zeiler2014visualizing,springenberg2014striving} have been proposed in the literature, which denoise the saliency map to gain higher sparsity. However, a drawback of these post-processing methods is the deviation from the original simple gradient map that alters the definition of the interpretation map.
For example, \citet{adebayo2018sanity} conducted a sanity check and found several interpretation methods such as Guided Propagation and Integrated Gradients lacked sensitivity to the model or the data-generating process. \citet{arun2021assessing} demonstrated similar findings on medical imaging applications. To illustrate the fidelity vs. sparsity trade-off in such post-processing schemes, we display an example in Fig.~\ref{fig:teaser}, where the simple gradient map looks to contain significant noise scattered over the background. A post-processing sparsification of the gradient map could raise the sparsity of the map; however, the sparsity comes at the cost of a considerable drop (more than 82\%) in the fidelity score. 

In this work, we propose a unified adversarial training (AT) framework to address the unstructured nature of standard simple-gradient maps, where we apply a norm-regularized adversarial training with a properly-designed norm function as an in-processing scheme to promote the sparsity structures in the simple gradient map. AT with standard $L_2$ and $L_\infty$-norm perturbations ~\cite{madry2017towards} has been successfully and widely utilized in the literature to enhance the adversarial robustness~\cite{szegedy2013intriguing} and interpretability~\cite{shah2021input} of deep neural networks. In our analysis, we extend the understanding of the influence of AT algorithms on the interpretability of a trained model, where we provide a convex duality framework to understand the influence of the norm function constrained in AT on the regularized norm of gradient maps and hence the interpretability of neural nets.

Specifically, we show a duality relation between the regularized norm of adversarial perturbations and the norm of input-based gradients. This duality relationship allows us to design norm-regularized adversarial training methods, which translates into the regularization of standard sparsity-inducing norms, e.g. the group norm \cite{yuan2006model} and the elastic net \cite{zou2005regularization}, of the simple gradient maps. As we show in this work, the special cases of our proposed framework can promote desired characteristics such as sparsity and connectedness in simple gradient maps. Importantly, unlike the sparsity-regularizing post-processing schemes, the gained sparsity properties of our proposed AT methods do not cost any loss in fidelity to the simple-gradient map, because we do not alter the definition of the interpretation map and only regularize the training process toward networks with more structured gradient maps.  


We conduct several numerical experiments on benchmark image datasets to validate the efficacy of our proposed AT-based methodology. The numerical results reveal the impact of our proposed norm-based AT methods on standard gradient-based maps, showing the enhanced sparsity and connectedness properties. We empirically analyze the performance of the trained neural nets in terms of several factors including interpretability, robustness, and stability. Our numerical results indicate the improvements in the mentioned structure-related factors offered by our designed norm-regularized AT methods. 
Finally, we utilize the shown duality relation to propose an interpretation harmonization scheme for aligning simple gradient maps with expert gaze maps, which performs satisfactorily in our numerical experiments. Our work's main contributions can be summarized as:

\begin{itemize}
\item We show a duality relation between the regularized norms of adversarial perturbations and gradient maps, and develop a unified AT framework for regularizing gradient maps.

\item We derive special cases of our proposed AT-based framework leading to the regularization of the elastic net and group-norm of gradient map.

\item We leverage the duality framework to propose an interpretation harmonization scheme 
for aligning gradient maps with human attention.

\item We provide numerical results showing the efficacy of the AT-based framework in improving the sparsity and stability of interpretation maps.

\end{itemize}
\label{sec:intro}

\section{Related Work}
\label{sec:relatedwork}
\textbf{Gradient-based Interpretation}. 
Using the gradient of the output of a deep neural network w.r.t. an input image is a widely-used approach to generate saliency maps~\cite{simonyan2013deep}. This method has been utilized in several related works and multiple variants of this approach are proposed in the literature, including SmoothGrad~\cite{smilkov2017smoothgrad}, Integrated Gradients~\citep{sundararajan2017axiomatic}, DeepLIFT~\citep{shrikumar2017learning}, and GradCAM~\citep{selvaraju2017grad}. The vanilla gradient-based maps without post-processing are often noisy and difficult to interpret. A group of methods are proposed to improve the visual quality by sparsifying the gradient-based maps: Guided Propagation~\citep{springenberg2014striving} removes the noise artifacts from saliency maps by suppressing negative activations in the back-propagation. Sparsified-SmoothGrad~\citep{levine2019certifiably} applies a sparsification to the saliency map to promote sparsity and robustness. MoreauGrad~\citep{zhang2023moreaugrad} generates saliency maps using an optimization problem with sparsity regularization. One drawback of the discussed methods is that they are all post-processing schemes, which may compromise the fidelity to the original simple-gradient map. On the other hand, our proposed framework is an in-processing scheme that remains faithful to the simple gradient map.

\noindent \textbf{Adversarial Training and Interpretability}.
Adversarial training (AT) involves training the networks using adversarial examples \cite{madry2017towards}. 
Fast Gradient Sign Method (FGSM)~\cite{goodfellow2014explaining} generates adversarial samples using a single iteration to create the adversarial training example for training the classifier. Projected gradient descent (PGD)~\cite{madry2017towards} uses several gradient-based steps to generate more effective adversarial examples in order to make the trained model more robust to norm-constrained adversarial attacks. 
While the primary objective of AT is to improve the trained model's robustness against adversarial attacks, several studies have shown that AT can also improve the visual quality of interpretation maps~\cite{ross2018improving,kim2019bridging,shah2021input}. Specifically, AT has been empirically shown to suppress irrelevant features~\cite{kim2019saliency}, and the standard $\ell_\infty$-norm-based AT will lead to higher sparsity and stability of the interpretation maps~\cite{chalasani2020concise}. Building upon and extending these findings, our work provides a unified duality framework that connects the perturbation norm constrained by the AT algorithm and the norm penalized for the interpretation map. Specifically, we discuss novel variants of AT methods that lead to the regularization of the elastic net and group norms of the interpretation maps which have not been studied in these related works.

\section{Preliminaries}
\label{sec:preliminaries}
\subsection{Gradient-based Saliency Maps}
Throughout this work, our analysis mainly focuses on the standard simple gradient as the gradient-based interpretation. We use $\mathbf{x} \in \mathbb{R}^d$ and $\mathbf{y} \in \mathbb{R}^C$ to denote the input and output of the neural network classifier. Note that $\mathbf{y}$ denotes the output of the neural net's post-softmax layer and contains the assigned likelihood for each of the $C$ classes in the classification task. The neural net function  $f_\theta: \mathbb{R}^d\rightarrow \mathbb{R}^C$ maps the input vector to the output vector, where $\theta$ denotes the parameters of the neural net. The simple gradient generates a saliency map by taking derivatives of the output of the neural network with respect to the input: 
\begin{equation*}
\label{eq:sg}
\mathrm{SG}(f_{\theta,c}, \mathbf{x}) := \nabla_\mathbf{x}f_{\theta,c}(\mathbf{x}).
\end{equation*}
Here, $c$ can be chosen as ground-truth label $y$ or the label $c$ with the maximum likelihood assigned by the classifier. In our analysis, we assume $\nabla_\mathbf{x} f_\theta{(\mathbf{x})}$ denotes the saliency map corresponding to the ground-truth label of sample $\mathbf{x}$.

\subsection{FGSM and PGD}
We consider two of the most widely used forms of adversarial training, FGSM and PGD, in this work. These methods first generate adversarial samples by calculating the gradient of the loss function with respect to the input data and then perturb the data by taking one or multiple small steps in the direction of the sign of the gradient. The perturbation is usually bounded by $l_\infty$-norm. The parameters of the network are then updated to minimize the loss with respect to these adversarial samples. We can use the following objective to summarize the optimization problem where $\widehat{P}$ is the empirical distribution of training samples
\begin{equation}
\label{eq:adv}
\min_{\theta} \;\mathbb{E}_{(\mathbf{x},y)\sim \widehat{P}}\Bigl[\max_{\Vert\delta\Vert_\infty \leq \epsilon}\; {\mathcal{L}(f_\theta(\mathbf{x}+\delta), y)}\Bigr].
\end{equation}
FGSM generates adversarial samples by taking one step in the direction of the sign of the gradient:
\begin{equation*}
    \mathbf{x}_{\mathrm{FGSM}}^* = \mathbf{x} + \epsilon \cdot\mathrm{sign}\bigl(\nabla_{\mathbf{x}}\mathcal{L}(f_\theta(\mathbf{x}), y)\bigr)
\end{equation*}
PGD, on the other hand, iteratively updates the samples with a stepsize $\alpha$:
\begin{equation*}
\mathbf{x}_{t+1} = \Pi_{\mathbf{x}+S} \bigl(\mathbf{x}_t + \alpha \mathrm{sign}(\nabla_{\mathbf{x}_t}\mathcal{L}(f_\theta(\mathbf{x}_t), y))\bigr),
\end{equation*}
where $S=\{\mathbf{z}:\: \Vert \mathbf{z}\Vert_\infty\le \epsilon\}$ denotes the $\epsilon$-radius $L_\infty$-ball.

\subsection{Fenchel Conjugate}
The Fenchel conjugate is an essential operation in convex analysis which has many applications to various computer vision algorithms. For a function $f: \mathcal{X} \rightarrow \mathbb{R}$, its Fenchel conjugate $f^\star : \mathcal{X} \rightarrow \mathbb{R}$ is defined as:
\begin{equation*}
f^\star(z) = \sup\bigl\{\langle x, z\rangle-f(x): x\in \mathcal{X}\bigr\},
\end{equation*}
where $\langle \cdot, \cdot\rangle$ denotes the inner product, and $\sup$ represents the supremum. Note that for a convex $f$, $f$ will be the Fenchel conjugate of $f^\star$. Furthermore, for every function $f$, the Fenchel conjugate $f^\star$ is guaranteed to be convex.

\section{An Adversarial Training  Methodology for Regularizing Interpretation Maps}
\label{sec:method}
In this section, we show a duality relation between the input gradient and the perturbation vectors, which is based on the linear approximation of the loss function $\ell\bigl(f(\mathbf{x}+{\delta}),y\bigr)$ around data point $\mathbf{x}$. In the approximation, the first-order term $\langle \nabla_\mathbf{x} \ell\bigl(f(\mathbf{x}),y\bigr) , {\delta}\rangle$ is the inner product of the perturbation vector and the input-based gradient of the loss function. Therefore, as will be explained in the section, to penalize a norm of the saliency map in the adversarial training process, we propose to constrain the dual norm of the perturbation vector. We summarize three variants of the norm constraint on the perturbation and its effect on the simple gradient saliency maps in Table~\ref{tab:overview}.

\begin{table}[tp]
\centering
\renewcommand\arraystretch{1.6}
    \caption{Summary of regularized norm $h$ on perturbation $\delta$ and the resulting regularized function $h^\star$ of the input gradient $\nabla_\mathbf{x}\mathcal{L}(f_\theta(\mathbf{x}), y)$. Detailed description can be found in Section~\ref{special}.}
\begin{center}
\vspace{-0.5cm}
\label{tab:overview}
\footnotesize
\begin{tabular}{c|c}
\toprule
$h(\delta)$&$h^\star(\nabla_\mathbf{x}\mathcal{L}(f_\theta(\mathbf{x}), y))$\\
\midrule
$\mathbb{I}(\Vert \delta \Vert_{\infty} \leq \epsilon)$&  $\epsilon\Vert \nabla_\mathbf{x}\mathcal{L}(f_\theta(\mathbf{x}), y)\Vert_1$\\
$\mathbb{I}(\Vert \delta \Vert_{2,\infty} \leq \epsilon)$&  $\epsilon\Vert \nabla_\mathbf{x}\mathcal{L}(f_\theta(\mathbf{x}), y) \Vert_{2,1}$\\
$\sum_i PQ_{\epsilon_1, \epsilon_2}(\delta_i)$& $\epsilon_1\Vert\nabla_\mathbf{x}\mathcal{L}(f_\theta(\mathbf{x}), y)\Vert_1 + \epsilon_2 \Vert\nabla_\mathbf{x}\mathcal{L}(f_\theta(\mathbf{x}), y)\Vert_2^2$\\
$\frac{1}{4\epsilon}\bigl\Vert\frac{1}{\mathbf{W}}\odot\delta\bigr\Vert_2^2$&$\epsilon\bigl\Vert\mathbf{W}\odot\nabla_\mathbf{x}\mathcal{L}(f_\theta(\mathbf{x}), y)\bigr\Vert_2^2$\\
\bottomrule
\end{tabular}
\end{center}
\vspace{-0.5cm}
\end{table}

\subsection{Rethinking the Impact of $\ell_\infty$-Norm-Based Adversarial Training on Saliency Maps}
We perform a first-order analysis by means of Fenchel conjugate to indicate how $L_\infty$-norm-based adversarial training could lead to higher sparsity in gradient-based maps. Under a small enough perturbation norm bound $\epsilon$ and twice-differentiable loss function, the objective of adversarial training in Eq.~\ref{eq:adv} can be approximated by first-order Taylor expansion as
\begin{align*}
\text{Eq.}~\ref{eq:adv} = \min_{\theta}\;&\mathbb{E}_{\mathbf{x},y\sim \widehat{P}}\Bigl[ \mathcal{L}(f_\theta(\mathbf{x}), y) \\
&+\max_{\Vert\delta\Vert_\infty \leq \epsilon}\:\bigl\{ \delta^T\nabla_\mathbf{x}{\mathcal{L}(f_\theta(\mathbf{x}), y)}\bigr\}\Bigr] + \mathcal{O}(\epsilon^2).
\end{align*}
Using the common cross-entropy loss or hinge loss, the derivative of the loss  with respect to the predicted logits at every training sample will be $\nabla_\mathbf{x}{\mathcal{L}(f_\theta,(\mathbf{x}), y)} = -f_{\theta,y}(\mathbf{x}) \nabla_\mathbf{x}{f_\theta(\mathbf{x})}$, so we can build the connection between the objective and the saliency map:
\begin{align*}
\text{Eq.}~\ref{eq:adv} = \min_{\theta}\mathbb{E}_{\mathbf{x},y\sim \widehat{P}}\Bigl[ &\mathcal{L}(f_{\theta, y}(\mathbf{x})) \\
+f_{\theta,y}(\mathbf{x})&\max_{\Vert\delta\Vert_\infty \leq \epsilon}\:\bigl\{ -\delta^T\nabla_\mathbf{x}{f_\theta(\mathbf{x})}\bigr\}\Bigr] + \mathcal{O}(\epsilon^2).
\end{align*}
Note we can leverage Fenchel conjugate operation to transform the maximization subproblem in the above problem into an unconstrained optimization: 
\begin{equation}
\label{eq:opt}
\max_{\Vert\delta\Vert_\infty \leq \epsilon} - \delta^T\nabla_\mathbf{x}{f_\theta(\mathbf{x})} = \max_\delta -\delta^T\nabla_\mathbf{\mathbf{x}}{f_\theta(\mathbf{x})} - \mathbb{I}(\Vert\delta\Vert_\infty \leq \epsilon),
\end{equation}
where $\mathbb{I}(\text{condition}) = 0$ if the condition holds and $+\infty$ otherwise. With the concept of Fenchel conjugate, the optimal value of the optimization problem defined in Eq.~\ref{eq:opt} is $\epsilon\Vert \nabla_\mathbf{x}{f_\theta(\mathbf{x})} \Vert_1$.  Hence, we successfully transform the original minimax optimization problem into a regularized optimization within an approximation error $\mathcal{O}(\epsilon^2)$: 
\begin{equation*}
\text{Eq.}~\ref{eq:adv} \approx \min_{\theta}\mathbb{E}_{\mathbf{x},y\sim\widehat{P}}\Bigl[ \mathcal{L}(f_\theta(\mathbf{x}), y)  + \epsilon f_{\theta,y}(\mathbf{x})\Vert \nabla_\mathbf{x}{f_\theta(\mathbf{x})} \Vert_1\Bigr].
\end{equation*}
Hence, we find that adversarial training with $L_\infty$-norm is approximately equivalent to applying $L_1$-norm regularization to the simple gradient saliency map, therefore promoting the sparsity of the saliency maps.

\subsection{Adversarial Training with Fenchel Conjugate Regularization: a Unified Approach}
Next, we consider adversarial training with a general norm-based regularization penalty. Here our goal is to promote a structured saliency map through a regularization function $h^\star(\mathbf{z})$. Note that for a convex $h$, the Fenchel conjugate of $h^\star(\mathbf{z})$ will be $h(\mathbf{x})$, Therefore, we consider the min-max optimization with the adversarial penalty function $-h(\mathbf{\delta})$:
\begin{equation*}
\label{eq:adv_fenchel}
\min_{\theta} \mathbb{E}_{\mathbf{x},y\sim\widehat{P}}\Bigl[\max_{\delta} \; {\mathcal{L}(f_\theta(\mathbf{x}+\delta), y)}-h(\delta)\Bigr].
\end{equation*}
Following our analysis for the $L_\infty$-norm-based adversarial training, 
we define the first-order approximate loss:
\begin{equation*}
\widehat{\mathcal{L}}(f_\theta(\mathbf{x}), y, \delta) :=  \mathcal{L}(f_\theta(\mathbf{x}), y) + \delta^T\nabla_x{\mathcal{L}(f_\theta(\mathbf{x}), y)},
\end{equation*}
To measure how well $\widehat{\mathcal{L}}(f_\theta(\mathbf{x}), y, \delta) $ approximates the adversarial loss, we observe the approximation error can be simply bounded for the class of $\lambda$-smooth functions.
\begin{definition} We call  $g:\mathbb{R}^d\rightarrow\mathbb{R}$ $\lambda$-smooth if for every $\mathbf{x}$ and $\mathbf{z}$: $\Vert\nabla g(\mathbf{x}) - \nabla g(\mathbf{z})\Vert_2 \leq \lambda \Vert \mathbf{x}-\mathbf{z}\Vert_2$.
\end{definition}
\begin{observation} 
\label{obs}
The approximation error of the first-order Taylor series expansion of adversarial loss under an $\epsilon$-$L_2$-norm bounded perturbation $\Vert\delta\Vert\le \epsilon$ can be bounded as 
$$\bigl\vert\widehat{\mathcal{L}}(f_\theta(\mathbf{x}), y, \delta) - {\mathcal{L}(f_\theta(\mathbf{x}+\delta), y)}\bigr\vert \leq \frac{\lambda \epsilon^2}{2}$$
\end{observation}
The observation shows $\widehat{\mathcal{L}}(f_\theta(\mathbf{x}), y, \delta) $ can be a good approximation of the adversarial loss under a small $\Vert\delta\Vert$. Combining the approximate loss with $h(\delta)$ regularization, we can derive the following equivalence by Fenchel conjugate:
\begin{proposition} Using Fenchel conjugate $h^\star$, we can reduce the maximization of the approximate adversarial loss as
\begin{equation}
\label{eq:fenchel}
\begin{split}
&\max_\delta\:\Bigl\{\widehat{\mathcal{L}}(f_\theta(\mathbf{x}), y, \delta) - h(\delta)\Bigr\} \\
=\:&\mathcal{L}(f_\theta(\mathbf{x}), y) +h^\star(\nabla_\mathbf{x}{\mathcal{L}\bigl(f_\theta(\mathbf{x}), y)}\bigr).
\end{split}
\end{equation}
\end{proposition}
 In a scenario that the neural network completely fits the training data, we have that $\nabla_\mathbf{x}{\mathcal{L}(f_\theta(\mathbf{x}), y)} = -f_{\theta,y}(\mathbf{x})\nabla_x{f_\theta(\mathbf{x})}\approx -\nabla_x{f_\theta(\mathbf{x})}$. Therefore, to enforce the gradient map $\nabla_x{f_\theta(\mathbf{x})}$ to satisfy a bounded penalty function $h^\star$, we can conduct adversarial training with an additive penalty term $-h(\delta)$. The optimal perturbation $\delta^*$ can be derived via the standard gradient ascent algorithm to maximize the regularized adversarial loss or be approximated with the analytic solution to the approximate optimization problem. 

\subsection{Special Cases of the Fenchel Conjugate-based Duality Framework}
\label{special}
Leveraging the duality framework in the previous section, we derive regularization penalty terms to penalize other sparsity norms different from $l_1$-norm and discuss their influence on the saliency maps.  

\noindent \textbf{$L_{2,1}$-group-norm penalty.} The $L_{2,1}$-group-norm of $\mathbf{x} \in \mathbb{R}^d$ for disjoint variable subsets $S_1, \cdots, S_t \subseteq  \{1, \cdots, d\}$ is defined as the special case $p=1$ of the following definition:
\begin{equation*}
 \Vert \mathbf{x}\Vert_{2,p} := \Vert[\Vert\mathbf{x}_{S_1}\Vert_2,\cdots,\Vert\mathbf{x}_{S_t}\Vert_2]\Vert_p.
\end{equation*}
To apply the duality framework to the group norm (with coefficient $\epsilon>0$), we derive its Fenchel conjugate as follows.
\begin{proposition}
\label{prop:2}
The Fenchel conjugate of the above $L_{2,1}$-group-norm is $h(\mathbf{z}) = \mathbb{I}(\Vert \mathbf{z}\Vert_{2, \infty} \leq \epsilon)$.
\end{proposition}

Different from $L_1$-norm inducing pixel-level sparsity, $L_{2,1}$-group-norm will induce sparsity of the patch basis. Under this regularization, the more influential patches are expected to be highlighted in the saliency map.
This effect could further lead to a connected saliency map, which will be desired when adjacent pixels belonging have similar effects on the prediction of the classifier. 


\noindent \textbf{Elastic net penalty.}
The elastic net penalty $h^\star(\mathbf{x})$ linearly combines the $L_1$ and $L^2_2$ penalty terms:
\begin{equation*}
h^\star(\mathbf{x}) = \epsilon_1\Vert\mathbf{x}\Vert_1 + \epsilon_2 \Vert\mathbf{x}\Vert_2^2, \epsilon_1>0,\epsilon_2 > 0.
\end{equation*}\vspace{-3mm}
\begin{proposition} 
\label{prop:3}
The Fenchel conjugate of the above elastic net penalty $h^\star$ is $
h(\mathbf{z}) = \sum_i PQ_{\epsilon_1, \epsilon_2}(\mathbf{z}_i)$ 
where $PQ_{\epsilon_1, \epsilon_2}$ is the piece-wise quadratic function defined as:
\begin{equation*}
PQ_{\epsilon_1, \epsilon_2}(\mathbf{z}_i)=
\begin{cases}
 \frac{1}{4\epsilon_2}(\mathbf{z}_i - \epsilon_1)^2 &  \text{\rm if}\;\; \epsilon_1< \mathbf{z}_i\\
\;\:\qquad 0& \text{\rm if}\;\;-\epsilon_1 \leq \mathbf{z}_i \leq \epsilon_1\\
 \frac{1}{4\epsilon_2}(\mathbf{z}_i + \epsilon_1)^2 & \text{\rm if}\;\;\mathbf{z}_i < -\epsilon_1.
\end{cases}
\end{equation*}
\end{proposition}
The linear combination of the $L_1$-norm and $L_2^2$-norm penalty terms in the elastic net results in a strongly convex regularization function, which is known to promote the stability of the sparsity pattern. Therefore, one can expect that the elastic net-regularized maps could possess higher robustness and stability than the $L_1$-norm regularized maps.

\noindent \textbf{Interpretation harmonization.}
The purpose of interpretation harmonization is to align the saliency map generated by the neural network with some reference attention map. The reference attention map can be manually labeled by domain experts, or based on some physical principle behind the classification task. Through interpretation harmonization, the network could be enabled to analyze the phenomenon in a similar way as a human, which would help diminish the potential bias or shortcuts learned by the neural net~\cite{geirhos2020shortcut,fel2022harmonizing}.

To achieve this goal, we leverage $L_2$-norm regularization to diminish the importance scores of unrelated features. Specifically, assume $\mathbf{A}\in \mathbb{R}^d$ is the expert's attention map with non-negative entries. A greater value implies a more important feature. We define $\Tilde{\mathbf{A}}:= \max(\mathbf{A})  - \mathbf{A} + \sigma$, where $\sigma$ is small enough so that the ordinal relation is reverted and all elements are positive. To diminish the importance score of the background region, we use the penalty term:
\begin{equation*}
h^\star(\mathbf{x}) = \epsilon\bigl\Vert \sqrt{\Tilde{\mathbf{A}}}\odot\mathbf{x}\bigr\Vert_2^2, \epsilon>0
\end{equation*}
where $\odot$ denotes Hadamard product. 
By adding this regularization term, the importance score of unrelated features is pushed to $0$ while important features remain unaffected.
\begin{proposition}
\label{prop:4}
The Fenchel conjugate of the above weighted $L_2$-norm square regularization is:
\begin{equation*}
h(\mathbf{z}) = \frac{1}{4\epsilon}\bigl\Vert\frac{1}{\sqrt{\Tilde{\mathbf{A}}}}\odot\mathbf{z}\bigr\Vert_2^2.
\end{equation*}
\end{proposition}
The derivation requires the weight term $\Tilde{\mathbf{A}}$ to be positive. However, a positive weight only leads to the diminishing of the importance score of unrelated features. It does not explicitly promote larger scores for important features. Empirically, we have found that using $\delta^*$ in the opposite direction to the gradients for important pixels helps increase the attribution scores for important features and results in better alignment. Therefore, in our numerical experiment, we set $\Tilde{\mathbf{A}}= \frac{1}{2}\max(\mathbf{A})-\mathbf{A}$ and $\delta^*=2\epsilon \Tilde{\mathbf{A}}\odot\nabla_\mathbf{x}{\mathcal{L}(f_\theta(\mathbf{x}), y)}$.
\section{Experiments}
\label{sec:experiments}
We conducted comprehensive numerical experiments to evaluate the performance of our proposed adversarial training-based methods. Firstly, we evaluate saliency maps' accuracy on a synthesized dataset with ground truth interpretation. Next, we assessed the visual quality, sparsity, interpretability, robustness, and stability of the saliency maps on Imagenette. Finally, we demonstrated the effectiveness of our interpretation harmonization strategy on the CUB-GHA dataset. Our quantitative results include adversarial training with one normalized step and multi-step gradient ascent. The main text focuses on the results of one normalized step, while detailed training settings and complete numerical results can be found in the Appendix. 
\begin{figure}[tp!]
    \centering
    \includegraphics[width=1\linewidth]{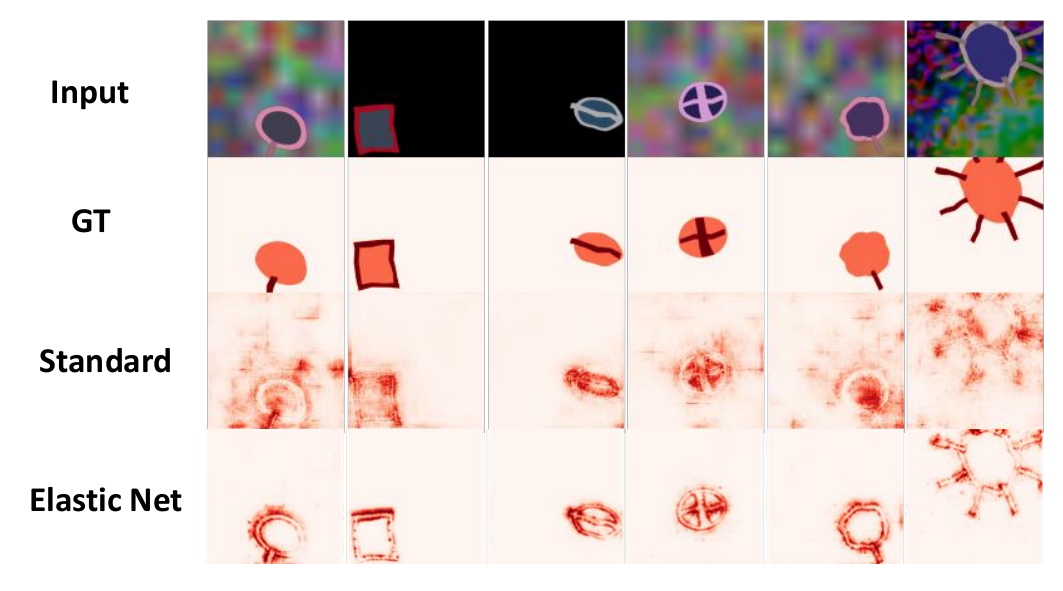}
    \caption{Qualitative results based on the synthesized dataset.}
    \label{fig:cell}
    \vspace{-5mm}
\end{figure}
\subsection{Results on synthesized dataset}
 \begin{table*}[tp]
\centering
\renewcommand\arraystretch{1}
    \caption{Comparison of five-band-scores and binary scores of standard training and adversarial training.}
\begin{center}
\label{tab:cell}
\vspace{-0.5cm}
\footnotesize
\begin{tabular}{c|cccc|cccc}
\toprule
\multirow{2}{4em}{Methods}&\multicolumn{4} {c|} {Five-band-scores}&\multicolumn{4} {c} {Binary scores}\\
\cline{2-9}
&Pixel\_acc ($\uparrow$) &Recall ($\uparrow$)&Precision ($\uparrow$)&FPR ($\downarrow$)&Pixel\_acc ($\uparrow$)&Recall ($\uparrow$)&Precision ($\uparrow$)&FPR ($\downarrow$)\\
\midrule
standard&89.81&9.86&32.58&1.21&88.66&44.47&16.22&10.22\\
$L_1$-norm&89.82&6.41&29.20&\textbf{0.76}&94.60&65.06&30.84&4.70\\
group-norm&89.84&7.54&27.35&0.91&94.27&63.78&29.77&4.99\\
elastic net&89.91&8.35&26.83&0.92&\textbf{94.63}&65.65&\textbf{31.29}&\textbf{4.66}\\
harmonization&\textbf{90.14}&\textbf{11.37}&\textbf{43.31}&1.14&93.82&\textbf{67.05}&28.15&5.54\\
\bottomrule
\end{tabular}
\end{center}
\vspace{-0.5cm}
\end{table*}

To illustrate how adversarial training can improve the model's interpretability, we first conducted experiments on a synthesized dataset with ground-truth for the saliency maps proposed by~\cite{tjoa2022quantifying}. There are three regions on the ground-truth of the synthesized dataset (Fig.~\ref{fig:cell}): background without any classification information (white region); localization information describes the location of an object (light red region); and distinguishing features crucial for classification (dark red area). We trained ResNet-34~\cite{he2016deep} with standard training and regularized adversarial training.

As the ground-truth was a ternary class, we reported the five-band-score reflects pixel accuracy, recall, precision, and the false positive rate (FPR) based on ternary classification. The results are listed in Table~\ref{tab:cell}, where the method's name corresponds to the penalty term $h^\star$ in Eq.~\ref{eq:fenchel}. The harmonization training achieved the best scores in terms of most of the metrics, showing the effectiveness of the proposed harmonization method. We also noticed that adversarial training-based methods consistently achieved higher accuracy and lower FPR compared with stranded training but led to lower recall and precision, because the adversarial training raised the sparsity of gradient maps. The importance scores of the localization area, which was less important compared with distinguishing features, were also diminished. Therefore, following~\cite{han2023impact}, we also converted the ternary classification task to a binary classification task by considering the background and localization information as one class. The results suggested that adversarial training could significantly improve the accuracy of saliency maps in detecting distinguishing features.

We visualized the saliency maps in Fig.~\ref{fig:cell} for qualitative evaluation (we only show results with elastic net regularization here and leave the rest in the Appendix). We discovered that the saliency map of standard training was activated more clearly in the region of the localization information than it was in the region of the distinguishing feature. 
In contrast, the saliency map with adversarial training was activated more significantly for the distinguishing features, especially on the boundaries. Moreover, the saliency maps looked more sparse and less noisy. 

\begin{figure}[tp!]
    \centering
\includegraphics[width=1\linewidth]{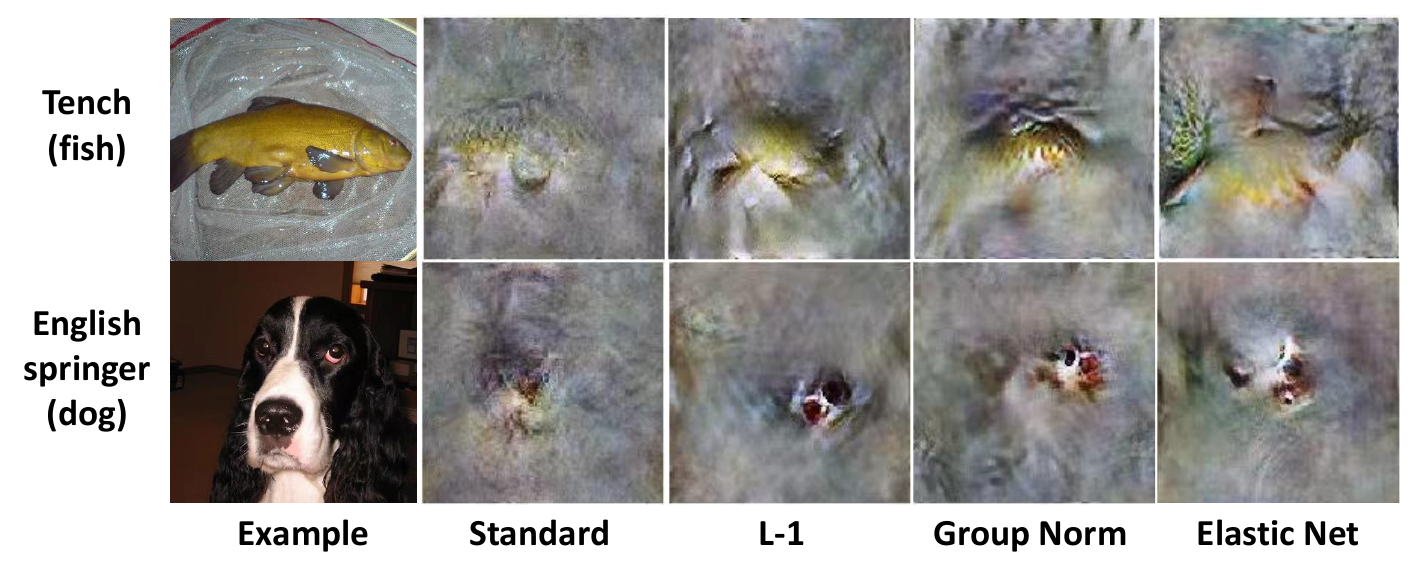}
    \caption{Images optimized for maximizing class-logit activations.}
    \label{fig:deep_dream}
\vspace{-0.5cm}
\end{figure}
\begin{figure*}[tp!]
    \centering
\includegraphics[width=1\linewidth]{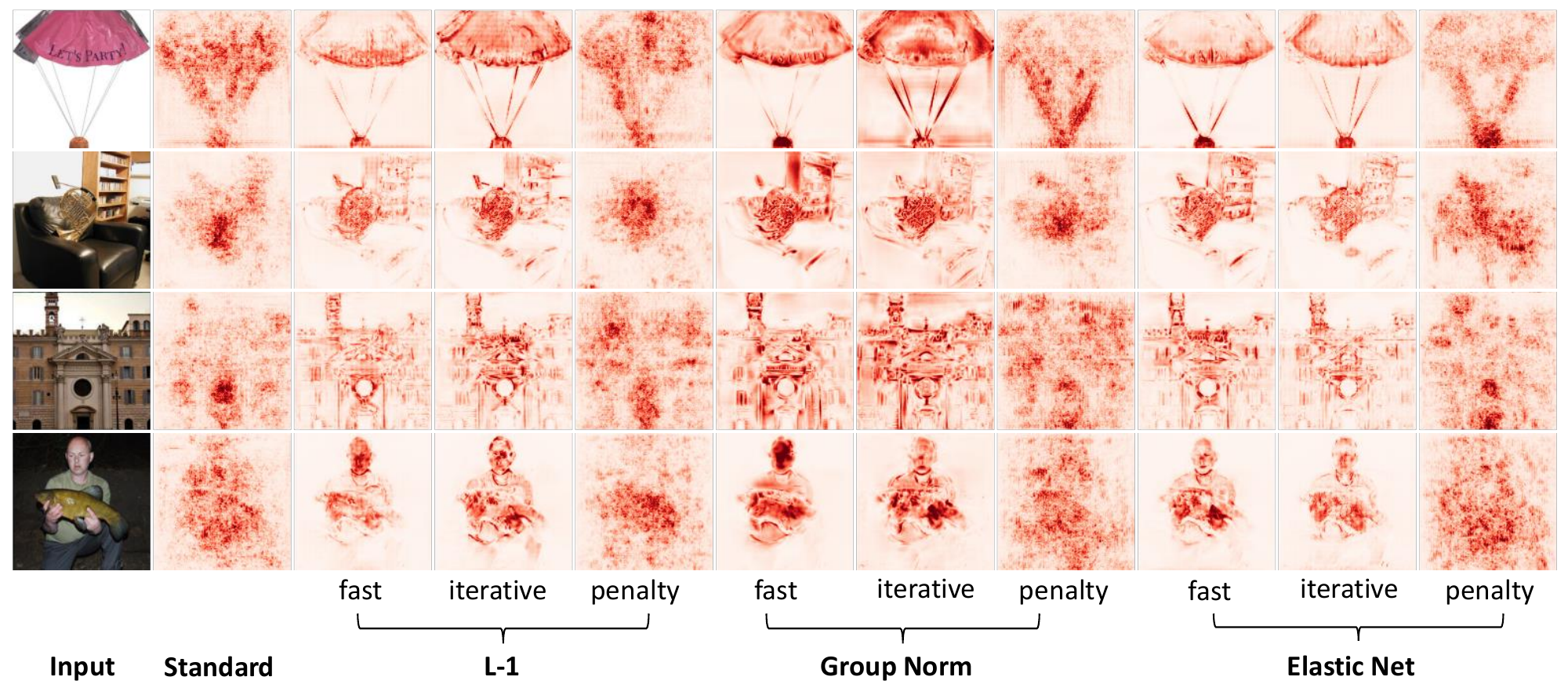}
\vspace{-0.7cm}
    \caption{Qualitative comparison of saliency maps generated by networks with different adversarial training protocols (fast, iterative) and standard training with additional regularization on input gradients (penalty).}
    \label{fig:training}
\vspace{-0.5cm}
\end{figure*}

\subsection{Results on ImageNette}
\label{sec:imagenette}
We performed numerical experiments on ImageNette, a ten-class subset of ImageNet~\cite{deng2009imagenet}, to examine the properties of the proposed adversarial training. Specifically, we studied the sparsity, interpretability, adversarial robustness, and stability of the saliency maps generated for  EfficientNet~\cite{tan2019efficientnet} classifiers with the proposed adversarial training methods. 

\noindent \textbf{Sparsity.} 
We trained the neural network models with different regularization coefficients and used the Gini Index~\cite{hurley2009comparing} as a measure of the sparsity of the saliency map. The results are shown in Table~\ref{tab:sparse}. All the proposed adversarial training-based methods led to more sparse saliency maps than standard training, with zero or only minor drops in clean accuracy. The $L_1$-norm-based method attained the highest sparsity. Also, the sparsity increased with the regularization coefficient $\epsilon$. The gradient maps visualized in Fig.~\ref{fig:training} suggest that standard training could lead to miscellaneous points in the background. On the other hand, the gradient maps with our proposed adversarial training methods seem to have higher visual quality. 

\noindent \textbf{Interpretability.} To show that the proposed training methods can improve the interpretability of the model, we calculated the DiffROAR score~\cite{shah2021input} for the methods. DiffROAR measures the difference in the predictive power of the dataset, where the top and bottom $k\%$ of the pixels are removed according to the importance score. A higher DiffROAR score shows the model captures the task-related features from the inputs. To this end, we measured the DiffROAR score for each method with $k$ from 10\% to 90\% in increments of 10\%. Each measurement was repeated with three different initializations, and the final DiffROAR score was the average of the 27 measurements. The results are shown in Table~\ref{tab:sparse}. Compared with standard training, the proposed methods can improve the model's interpretability, especially under the elastic net regularization, which achieved the highest DiffROAR score. 

We also visualized the learned feature for each model, using the optimization-based feature visualization technique of \textit{deep-dream}~\cite{olah2017feature}. The results (Fig.~\ref{fig:deep_dream}) also show that class-specific features identified with the proposed norm-regularized adversarial training methods look more meaningful to human perception.

\begin{figure*}[tp]
    \centering
\includegraphics[width=1\linewidth]{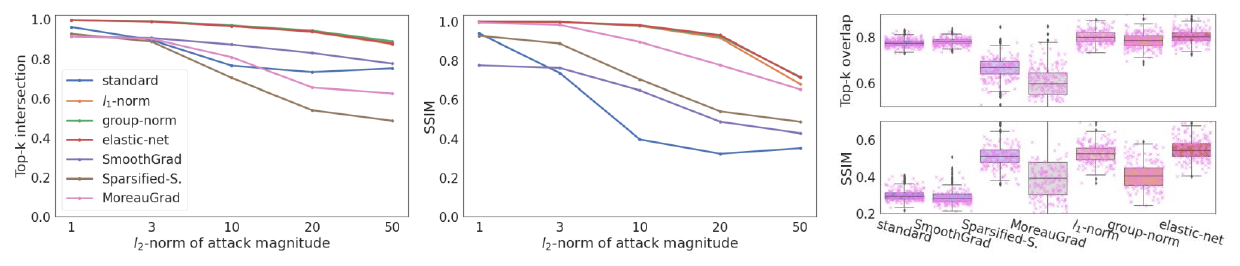}
\vspace{-0.7cm}
    \caption{Quantitative robustness and stability comparison. \textbf{Left}/\textbf{Middle}: Comparison of SSIM/top-k intersection of saliency maps before and after the attack. \textbf{Right}: Comparison of SSIM/top-k overlap of saliency maps generated by networks with different stochastic training.}
    \label{fig:robustness}
\end{figure*}

\noindent \textbf{Robustness.} 
To assess the robustness of interpretation maps under the standard and proposed training methods, we adopted an $L_2$-norm bounded interpretation attack method proposed by~\cite{levine2019certifiably}. We gradually increased the strength of the attack and measured the robustness through similarity measures of the saliency maps before and after the attack. We adopted two metrics as robustness measures: 1) top-k intersection ratio~\cite{ghorbani2019interpretation}, which is the ratio of pixels that remain salient after the interpretation attack, and 2) the structural similarity index measure (SSIM)~\cite{wang2004image}. We compared the proposed training methods with the baseline standard training and three post-processing-based baselines, SmoothGrad, Sparsified-SmoothGrad (Sparsified-S.), and MoreauGrad (Fig.~\ref{fig:robustness}). Standard training seemed vulnerable to minor attacks, even when combined with the Smooth-based post-processing. In contrast, the neural nets trained by the proposed methods displayed significantly higher robustness to the attacks. We also observed that the elastic net and group-norm-based training methods performed slightly more robustly than the $L_1$-norm-based method.
 \begin{table}[tp]
\centering
\renewcommand\arraystretch{1}
    \caption{Quantitative evaluation on ImageNette.}
\begin{center}
\label{tab:sparse}
\footnotesize
\vspace{-0.5cm}
\begin{tabular}{c|ccc}
\toprule
Methods & \makecell[c]{DiffROAR \\ (\%) ($\uparrow$)} & \makecell[c]{Gini \\ (\%) ($\uparrow$)} & \makecell[c]{Acc. \\ (\%) ($\uparrow$)}\\
\midrule
standard&1.81&46.53&89.04\\
\midrule
$L_1$-norm ($\epsilon$=0.01)&2.95&56.73&\textbf{90.46}\\
$L_1$-norm ($\epsilon$=0.05)&2.12&62.04&84.13\\
$L_1$-norm ($\epsilon$=0.10)&1.97&\textbf{65.53}&78.85\\
\midrule
group-norm ($\epsilon$=0.10)&2.57&50.07&86.83\\
group-norm ($\epsilon$=0.50)&1.32&49.15&87.57\\
group-norm ($\epsilon$=1.00)&0.40&52.70&83.80\\
\midrule
elastic net ($\epsilon_1$=0.01, $\epsilon_2$=0.01)&2.60&57.10&87.72\\
elastic net ($\epsilon_1$=0.01, $\epsilon_2$=0.05)&\textbf{3.24}&57.35&87.67\\
elastic net ($\epsilon_1$=0.01, $\epsilon_2$=0.10)&3.14&54.44&87.19\\
elastic net ($\epsilon_1$=0.05, $\epsilon_2$=0.01)&2.31&60.05&84.59\\
elastic net ($\epsilon_1$=0.05, $\epsilon_2$=0.05)&2.11&61.84&84.69\\
elastic net ($\epsilon_1$=0.05, $\epsilon_2$=0.10)&2.17&60.25&84.54\\
\bottomrule
\end{tabular}
\end{center}
\vspace{-0.7cm}
\end{table}

\noindent \textbf{Stability.} We evaluated the algorithmic stability as the magnitude of perturbation in the saliency maps under classifiers trained with different training data or random initialization. As pointed out by~\cite{woerl2023initialization}, standard training with different initializations can result in large discrepancies in saliency maps. We examined if the proposed training methods can alleviate this issue. To this end, we switched 10\% of the training data with the test data and initialized two networks with different random seeds. Then we trained these two networks with the same training process. We used SSIM and top-k overlap to measure the similarities of the generated saliency maps, where the top-k overlap was defined as the Dice score of the top-k masks. As shown in Fig.~\ref{fig:robustness}, adversarial training could significantly improve stability. It even brings more improvement than post-processing methods. Note for Sparsified-SmoothGrad, the high SSIM is due to the sparsity of the saliency maps. Through studying top-k overlap, the activated regions were actually different. Among the three variants of adversarial training, the elastic net could achieve the best stability, as the incorporation of $L^2_2$ term could increase the smoothness of the objective function, making the optimization process less sensitive to the perturbation of the training data. 

\noindent \textbf{Comparison of different training protocols.}
We also conducted experiments to see how different training protocols affect the numerical results. Specifically, we compared one-step optimization (fast), multiple-step optimization (iterative), and free adversarial training (free, in Appendix)~\cite{shafahi2019adversarial}. As a baseline, we also studied standard training with norm regularization directly applied to the objective function (penalty). We visualized a few saliency maps in Fig.~\ref{fig:training} for comparison. Standard training with regularization failed to generate structured saliency maps, showing the challenge of network optimization with regularization in the objective function, which further certified the superiority of adversarial training.  Visually, fast, and iterative training achieved comparable quality. However, for $L_1$-norm, interactive training would lose sparsity compared to fast training, but for the elastic net, the trend was the opposite. The Fenchel conjugate of elastic net regularization is differentiable everywhere due to the square term, making its optimization easier compared with $L_1$-norm.

\subsection{Harmonization with gaze maps}
We conducted experiments on the CUB-GHA~\cite{rong2021human} to verify the effectiveness of our interpretation harmonization strategy on real-world data. CUB-GHA is an extension of CUB~\cite{wah2011caltech}, which includes gaze maps collected from domain experts for bird category classification. The gaze map is a type of human attention map reflecting how human experts would conduct this task. We aim to harmonize the saliency maps with human attention so that the network could make the decision more similar to humans. 

\begin{figure}[tp!]
    \centering
\includegraphics[width=0.95\linewidth]{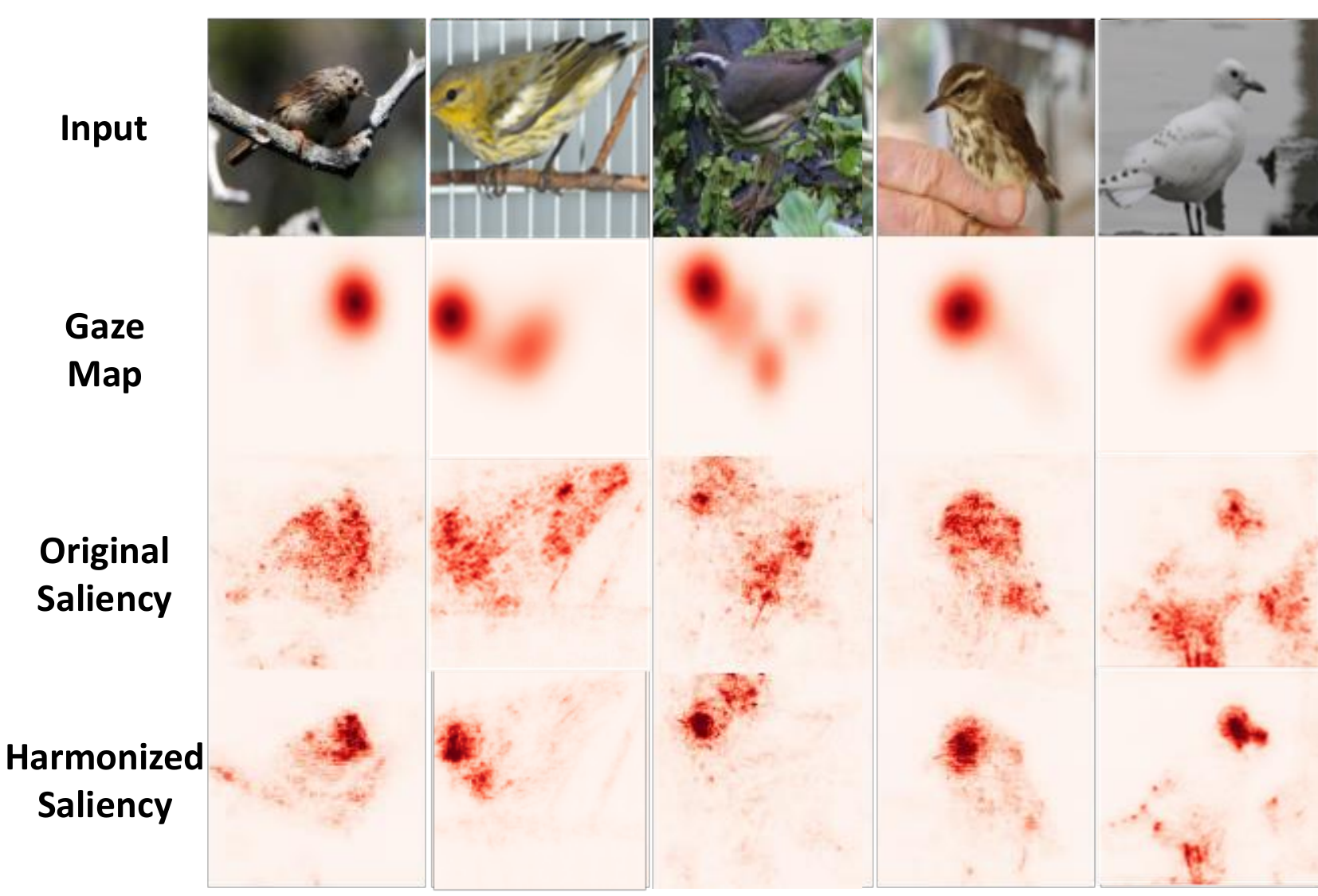}
    \caption{Qualitative results based on CUB-GHA dataset.}
    \label{fig:cub}
\vspace{-0.5cm}
\end{figure}

To do this, we normalized the gaze map and used the attention score as the weights for the $l_2$-norm perturbation during adversarial training. The results are shown in Fig.~\ref{fig:cub}. The original saliency maps highlighted the whole body of the bird, while the gaze maps were more focused on a specific part (e.g. head) of the bird. After harmonization, the saliency maps showed better alignment with the gaze maps. We also calculated the top-k overlap between saliency maps and gaze maps (Table~\ref{tab:harmonize}). The results showed an increasing trend of the overlap as we raised the coefficient $\epsilon$, while the trade-off of accuracy was in a reasonable range. 

 \begin{table}[tp]
\centering
\renewcommand\arraystretch{1}
    \caption{Top-k overlap (\%) between gaze map and saliency map on CUB-GHA dataset. }
\begin{center}
\vspace{-0.5cm}
\footnotesize
\label{tab:harmonize}
\begin{tabular}{c|ccccc}
\toprule
$\epsilon$ & 0.0 & 0.1 & 0.5 & 1.0 & 5.0 \\
\midrule
top 5\% overlap& 33.42&35.07&38.74&42.63&50.62\\
top 10\% overlap& 41.81&42.99&45.60&47
06&52.60\\
\midrule
accuracy  (\%)&67.58 & 67.50 & 66.56&66.65&64.99\\
\bottomrule
\end{tabular}
\end{center}
\end{table}

\section{Conclusion}
\label{sec:conclusion}
In this paper, we introduced a duality framework to analyze the impact of norm-regularized adversarial training on the gradient-based saliency maps of a trained neural net classifier. Leveraging this duality framework, we proposed several variants of norm-regularized adversarial training, designed to promote certain structures in the gradient-based interpretation maps, including sparsity, group sparsity, and consistency with human attention. We provided experimental results on several benchmark datasets to validate the effectiveness of our proposed methods, which demonstrated that properly designed adversarial training methods can enhance model interpretability as well as the stability and robustness of the gradient-based maps. An interesting future direction is to develop a similar adversarial training-based regularization framework for structured interpretation maps according to integrated gradients, DeepLIFT, and Grad-Cam. Another relevant future direction is to explore other forms of penalty terms in order to expand the methodology to domains beyond computer vision. \vspace{1mm}

\noindent\textbf{Acknowledgments.}
The work of Farzan Farnia is partially supported by a grant from the
Research Grants Council of the Hong Kong Special Administrative Region, China, Project 14209920, and is partially supported by a CUHK Direct Research Grant with CUHK Project No. 4055164. 

\newpage
{
    \small
    \bibliographystyle{ieeenat_fullname}
    \bibliography{main}
}

\clearpage
\setcounter{page}{1}
\maketitlesupplementary
\appendix
In this supplementary material, we first gave the proofs for some of the propositions (Sec.~\ref{proof}) and implementation details for the experiments in the main text (Sec.~\ref{implementation}). Then we added additional experimental results to illustrate the efficacy of our proposed method further. For the synthesized dataset, we gave a complete visualization of the saliency maps generated with adversarial training (Sec.~\ref{sec:add_synthesized}). For ImageNette, we compared with additive Gaussian noise during training (Sec.~\ref{sec:img_1}), free adversarial training (Sec.~\ref{sec:img_2}), and adversarial attacks with momentum (Sec.~\ref{sec:mifgsm}). We further showed that our in-processing scheme can better preserve the fidelity of the interpretation compared with post-processing methods (Sec.~\ref{sec:fidelity}) while integrating both methods could produce visually coherent maps (Sec.~\ref{sec:img_3}). In addition, we conducted sanity checks to show our method guarantees high fidelity (Sec.~\ref{sec:img_4}). Finally, we visualized the results of interpretation attacks to further illustrate the robustness of our method (Sec.~\ref{sec:img_5}). For the CUB-GHA dataset, we compared our interpretation harmonization strategy with vanilla $l_2$-norm-regularized adversarial training (Sec.~\ref{sec:img_6}).  

\section{Proofs}
\label{proof}
In this section, we gave the proofs for some of the observations and propositions.
\subsection{Proof of Observation~\ref{obs}}
According to the Taylor's theorem, we have 
\begin{equation*}
\begin{split}
\bigl\vert\widehat{\mathcal{L}}(f_\theta(\mathbf{x}), y, \delta) - &{\mathcal{L}(f_\theta(\mathbf{x}+\delta), y)}\bigr\vert \\= &\vert\frac{1}{2}\delta^T\mathbf{H}_{\mathcal{L}(f_\theta(\mathbf{x}), y))}(\mathbf{x}+\gamma\delta)\delta\vert\\
\leq &\frac{1}{2} \Vert\delta\Vert^2 \Vert\mathbf{H}_{\mathcal{L}(f_\theta(\mathbf{x}), y))}(\mathbf{x}+\gamma\delta)\Vert\\
\leq & \frac{1}{2}\lambda \epsilon^2.
\end{split}
\end{equation*}
Where $\mathbf{H}_{\mathcal{L}(f_\theta(\mathbf{x}), y))}(\cdot)$ is the Hessian matrix of the $\mathcal{L}(f_\theta(\mathbf{x}), y))$ w.r.t $\mathbf{x}$ and $\gamma \in [0,1]$. The last in-equation holds because we assume $f_\theta(\mathbf{x})$ is $\lambda$-smooth and $\delta$ is $\epsilon$-bounded.

\subsection{Proof of Proposition~\ref{prop:2}}
\begin{equation*}
\begin{split}
h(\mathbf{z})&=\sup_\mathbf{x}\{\mathbf{x}^T\mathbf{z}-\epsilon\Vert\mathbf{x}\Vert_{2,1}\}\\
&= \sum_{j=1}^t\sup_{\mathbf{x}_{S_j}}\{\mathbf{x}_{S_j}^T\mathbf{z}_{S_j}-\epsilon\Vert\mathbf{x}_{S_j}\Vert\}\\
& = \sum_{j=1}^t\sup_{t\geq 0}\sup_{\Vert\mathbf{x}_{S_j}\Vert=t}\{\Vert\mathbf{z}_{S_j}\Vert t-\epsilon t\}\\
& =\begin{cases}
0 &\text{\rm if}\;\; \Vert\mathbf{z}_{S_j}\Vert \leq \epsilon, \forall j\\
+\infty &  \text{\rm else}\\
\end{cases}\\
&= \mathbb{I}(\Vert \mathbf{z}\Vert_{2, \infty} \leq \epsilon).
\end{split}
\end{equation*}

\subsection{Proof of Proposition~\ref{prop:3}}
\begin{equation*}
\begin{split}
h(\mathbf{z})&=\sup_\mathbf{x}\{\mathbf{x}^T\mathbf{z}-\epsilon_1\Vert\mathbf{x}\Vert_{1} - \epsilon_2\Vert\mathbf{x}\Vert_{2}^2\}\\
&= \sum_{i=1}^n\sup_{\mathbf{x}_i}\{\mathbf{x}_i\mathbf{z}_i - \epsilon_1\vert\mathbf{x}_i\vert-\epsilon_2\mathbf{x}_i^2\}\\
& = \sum_{i=1}^n PQ_{\epsilon_1, \epsilon_2}(\mathbf{z}_i),\\
\end{split}
\end{equation*}
where
\begin{equation*}
PQ_{\epsilon_1, \epsilon_2}(z)=
\begin{cases}
 \frac{1}{4\epsilon_2}(z - \epsilon_1)^2 &  \text{\rm if}\;\; \epsilon_1< z\\
\;\:\qquad 0& \text{\rm if}\;\;-\epsilon_1 \leq z \leq \epsilon_1\\
 \frac{1}{4\epsilon_2}(z + \epsilon_1)^2 & \text{\rm if}\;\;z < -\epsilon_1.
\end{cases}
\end{equation*}
\subsection{Proof of Proposition~\ref{prop:4}}
This proposition holds by reusing the proof of Proposition~\ref{prop:3} with $\epsilon_1 = 0$ and setting different $\epsilon_2$ for every element of $\mathbf{z}$.

\section{Implementation details}
\label{implementation}
In this section, we introduced the implementation details for the experiments. All experiments were performed in PyTorch 1.12.1 using one Nvidia GeForce RTX 2080 GPU.
\label{sec:implementation}
\subsection{Experimental setup for synthesized dataset}
The synthesized dataset comprised 10 classes. We generated 4096 images for training and 1024 images for testing. All images were of the size $512\times 512$. The synthesized dataset formed a hierarchy with a circular, rectangular, and tail category. Each sample exhibited one among the three types (dark, blurred, and noisy) of random background. We used ResNet-34~\cite{he2016deep} pre-trained on ImageNet as the classification network and trained the network with the Adam optimizer. The experiments used a batch size of 16, an initial learning rate of 0.001, a momentum of (0.5, 0.999), and
a weight decay of $10^{-5}$. As the task is very simple and easy to over-fit, we set the number of training epochs to be 10.

\subsection{Setup for training ImageNette}
\label{setup:imagenet}
ImageNette is a smaller subset of 10 easily classified classes from Imagenet. The dataset contains 9469 images for training and 3925 images for testing. To train a classification network, we re-sampled all images into the size of $224 \times 224$ and then normalized the images to have zero mean and unit standard deviation. We used Efficientnet-B0~\cite{tan2019efficientnet} as the classification network, which was trained by Adam optimizer with an initial learning rate of $3\times10^{-4}$, a momentum of 0.999, and
a weight decay of $10^{-4}$. The experiments use a batch size of 16. We set the number of training epochs to be 200. 
\subsection{Details of one-step optimization for adversarial training.}
To speed up the adversarial training, we utilized the analytical solution to the approximate optimization problem given in Eq.~\ref{eq:fenchel} as the adversarial perturbation, which could be derived without iteration. According to Observation~\ref{obs}, this solution approximates the optimal solution well. Specifically, we gave the formula for the approximate solution:
\begin{itemize}
\item \textbf{$L_1$-norm}: $$\delta^* = \epsilon\text{sgn}(\nabla_\mathbf{x}{\mathcal{L}(f_\theta(\mathbf{x}), y)}).$$
\item \textbf{$L_{2,1}$-group-norm}: $$\delta^* = \epsilon[\frac{\nabla_{\mathbf{x}_{S_1}}{\mathcal{L}(f_\theta(\mathbf{x}), y)}}{\Vert\nabla_{\mathbf{x}_{S_1}}{\mathcal{L}(f_\theta(\mathbf{x}), y)}\Vert}, \cdots, \frac{\nabla_{\mathbf{x}_{S_t}}{\mathcal{L}(f_\theta(\mathbf{x}), y)}}{\Vert\nabla_{\mathbf{x}_{S_t}}{\mathcal{L}(f_\theta(\mathbf{x}), y)}\Vert}].$$
\item \textbf{elastic net}: $$\delta^* = \epsilon_1\text{sgn}(\nabla_\mathbf{x}{\mathcal{L}(f_\theta(\mathbf{x}), y)}) + 2\epsilon_2\nabla_\mathbf{x}{\mathcal{L}(f_\theta(\mathbf{x}), y)}.$$
\end{itemize}
We noted that for $L_1$-norm regularization, the analytical solution to the approximate optimization problem gives exactly FGSM.

\subsection{Details of iterative adversarial training.}
We conducted gradient ascend for iterative adversarial training. The number of steps was set to be 7, and the step size is 0.3. We normalized the norm of the gradients to stabilize the training. For $L_1$-norm, we followed the original training of PGD to utilize the sign of the gradients. For elastic-net and group-norm we normalized the gradients so it has unit $L_2$-norm. Moreover, to facilitate convergence, we initialized the perturbation with the analytical solution to the approximate optimization problem given in Eq.~\ref{eq:fenchel} of the main paper.

\subsection{Setup for calculating DiffROAR}
In the main paper, we measured the DiffROAR scores (in Sec.~\ref{sec:imagenette}). DiffROAR is the difference in the predictive power of datasets, with the top-k\% and bottom-k\% of pixels removed by ordering the feature importance of the model. To measure this score, we re-trained the ResNet-34 model for top-k\% and bottom-k\% removed datasets with 150 epochs. The training setup was the same as in Sec.~\ref{setup:imagenet}. The differences in accuracy for top-k\% and bottom-k\% in the test set were presented in the main paper. 

\subsection{Setup for comparing the robustness.}
To verify the robustness of the interpretation methods, we followed~\cite{levine2019certifiably} to use $L_2$ attack on top-k overlap. Here $k$ was selected as 40\% of all the pixels. For SmoothGrad and Sparsified-SmoothGrad, we used the sample size of 64 and applied fp16 to attack, due to memory constraints. For calculating the top-k intersection metric, we set $k$ to be 40\% of all the pixels with non-zero attributions.  The evaluation metrics were calculated based on 500 images randomly selected from the test set.
\begin{figure}[tp!]
    \centering
    \includegraphics[width=1\linewidth]{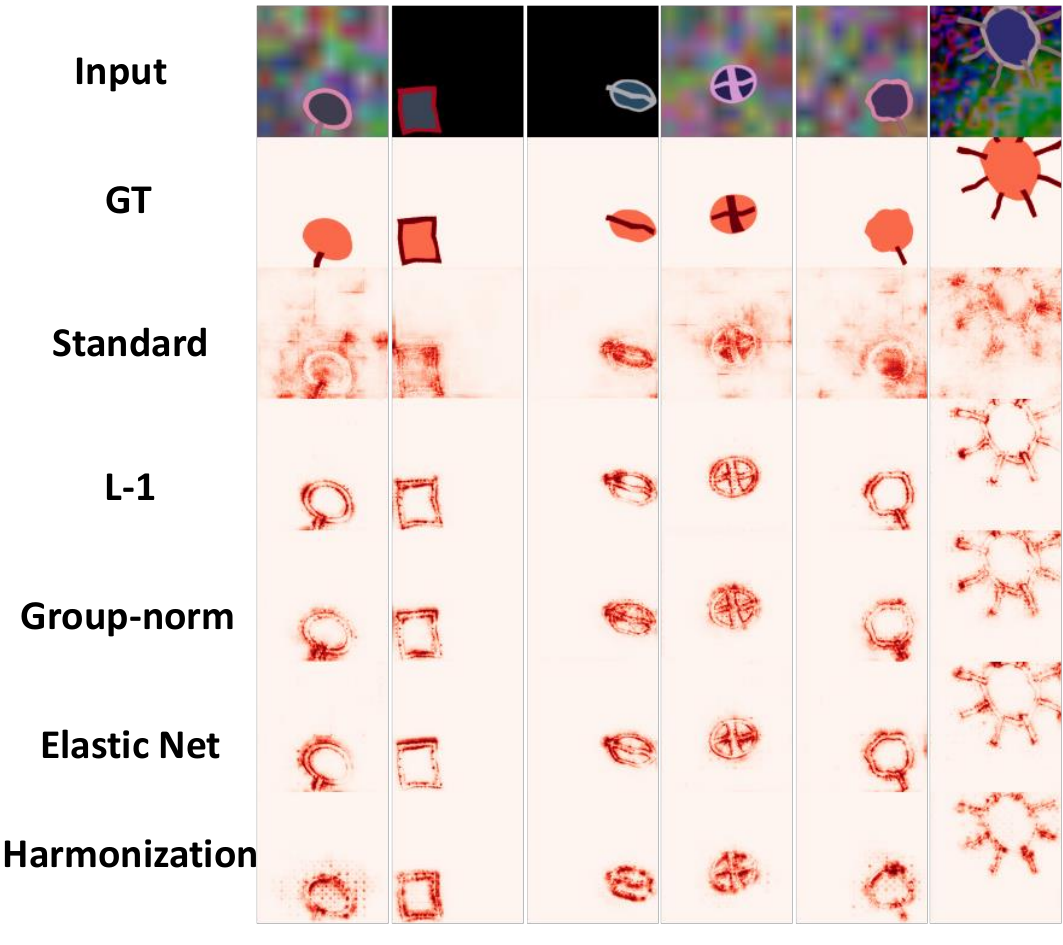}
    \caption{Qualitative results based on the synthesized dataset.}
    \label{fig:cell_supp}
\end{figure}

\subsection{Setup for comparing the stability.}
To examine the stability of the interpretation methods, we trained two networks with the same training process but with different initializations. Moreover, to count for the influence of training data, we randomly removed 1000 images from the training set and replenished the same amount of images from the test set. The metrics were calculated based on 500 images randomly selected from the common test set of the two runs. For calculating the top-k intersection metric, we set $k$ to be 40\% of all the pixels with non-zero attributions.
\begin{figure}[tp!]
    \centering
    \includegraphics[width=1\linewidth]{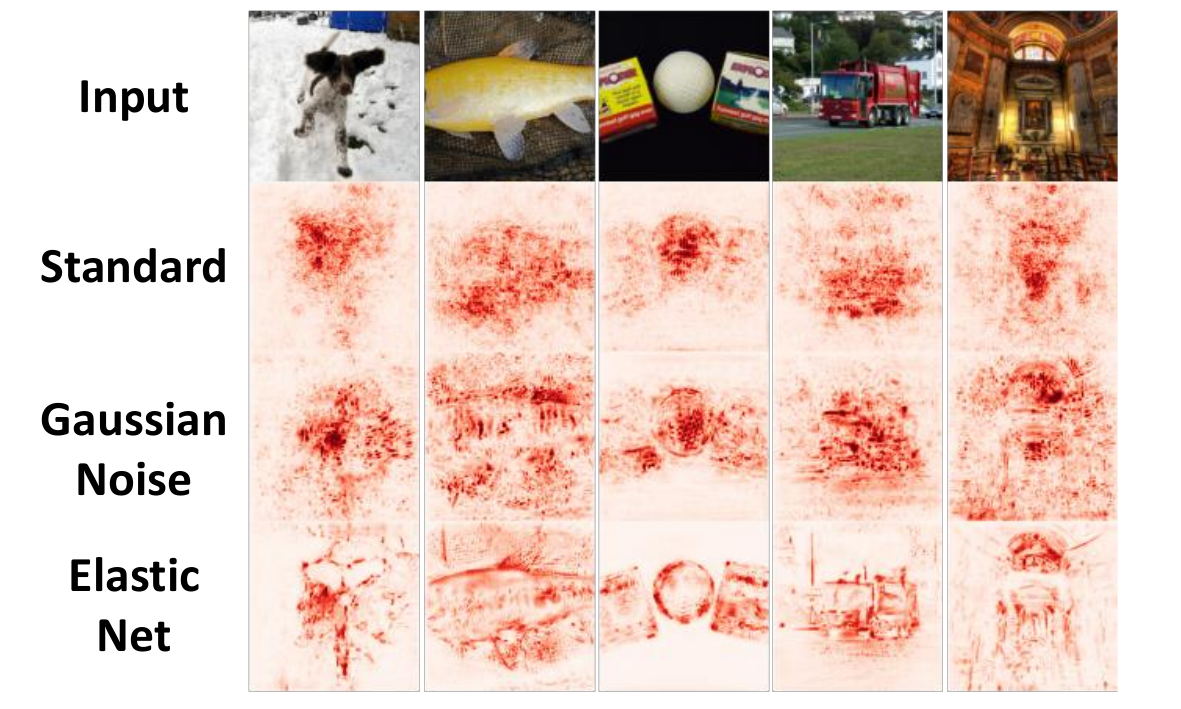}
    \caption{Adding noise during training vs. adversarial training.}
    \label{fig:smooth}
    \vspace{-3mm}
\end{figure}

\subsection{Experimental setup for CUB-GHA}
CUB is a dataset for bird category classification. It is composed of 5990 training images and 5790 test images of 200 kinds of birds. CUB-GHA further collected human gaze data and Gaussian filter on fixation points to generate human attention maps. We resized the image to be $600 \times 600$ and randomly cropped it to be $448 \times 448$ before plugging it into an Efficientnet-B0 for classification. The remaining settings remained the same as in Sec.~\ref{setup:imagenet}. For adversarial training, the perturbation was calculated as an element-wise multiplication of the processed attention map and the gradients. 
We further normalized the gradients with $L_2$-norm to stabilize the training process:
\begin{equation*}
\delta^* = (\epsilon\max(A) - 2\epsilon A)\odot \frac{\nabla_\mathbf{x}{\mathcal{L}(f_\theta(\mathbf{x}), y)}}{\Vert \nabla_\mathbf{x}{\mathcal{L}(f_\theta(\mathbf{x}), y)} \Vert}.
\end{equation*}

\section{Additional results}
\label{sec:add_results}
\subsection{Results on synthesized dataset}
\subsubsection{Visualization of different norm-regularization} 
\label{sec:add_synthesized}
In addition to the saliency map of the elastic net given in the main text, we visualized the saliency maps corresponding to other norm regularizations in Fig.~\ref{fig:cell_supp}. As the task is very simple and the images are very clean, different norm regularizations do not lead to quantitatively different visualizations. However, they all show significant improvement over standard training.

\subsection{Results on ImageNette}
\subsubsection{Comparison with adding noise during training}
\label{sec:img_1}
Adding random noise during training also has the effect of denoising saliency maps and producing saliency maps of high visual quality~\cite{smilkov2017smoothgrad}. One may think the denoising effects of adversarial training may come from adding noise during training. To illustrate it, we compared adversarial training with adding Gaussian noise during training. The results are shown in Fig.~\ref{fig:smooth}. Although training with additive Gaussian noise resulted in less noisy saliency maps compared with standard training, it still showed much worse visual quality compared with adversarial training. This shows adversarial perturbation indeed has effects on the sparseness of the saliency maps.

\subsubsection{Results of free adversarial training}
\label{sec:img_2}
\begin{figure*}[tp!]
    \centering
    \includegraphics[width=1\linewidth]{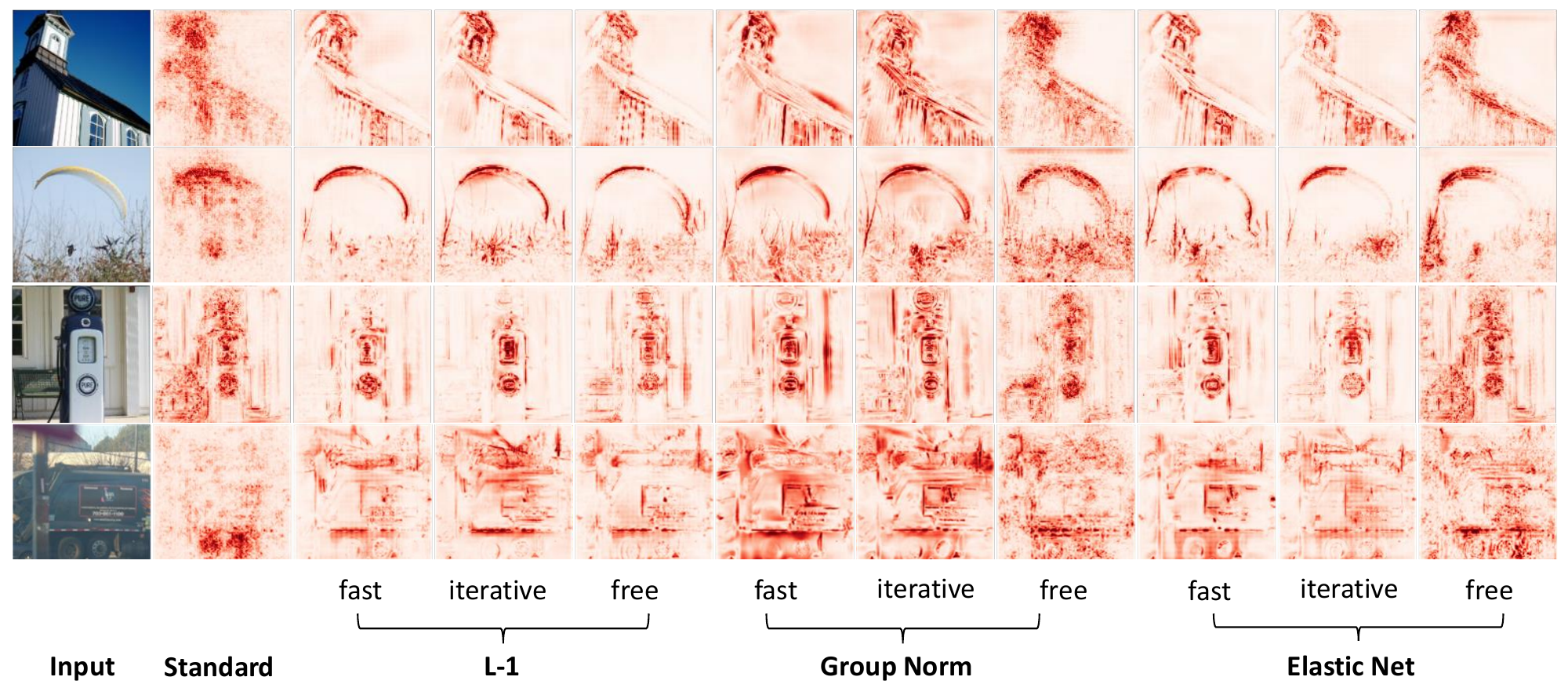}
    \caption{Qualitative comparison of saliency maps generated by networks with different adversarial training protocols (fast, iterative, free).}
    \label{fig:train_supp}
    \vspace{-5mm}
\end{figure*}

In addition to one-step optimization and iterative optimization, we also conducted experiments with free adversarial training~\cite{shafahi2019adversarial}, which optimize the network parameter and the adversarial perturbation simultaneously, so as to expedite the training. The results are shown in Fig.~\ref{fig:train_supp}. As the results show, free adversarial training can also be used to conduct norm-regularized adversarial training. However, the visual quality is slightly lower than that of fast and iterative training. Hence, for interpretation purposes, fast or iterative training is recommended.

\begin{figure}[tp!]
    \centering
    \includegraphics[width=1\linewidth]{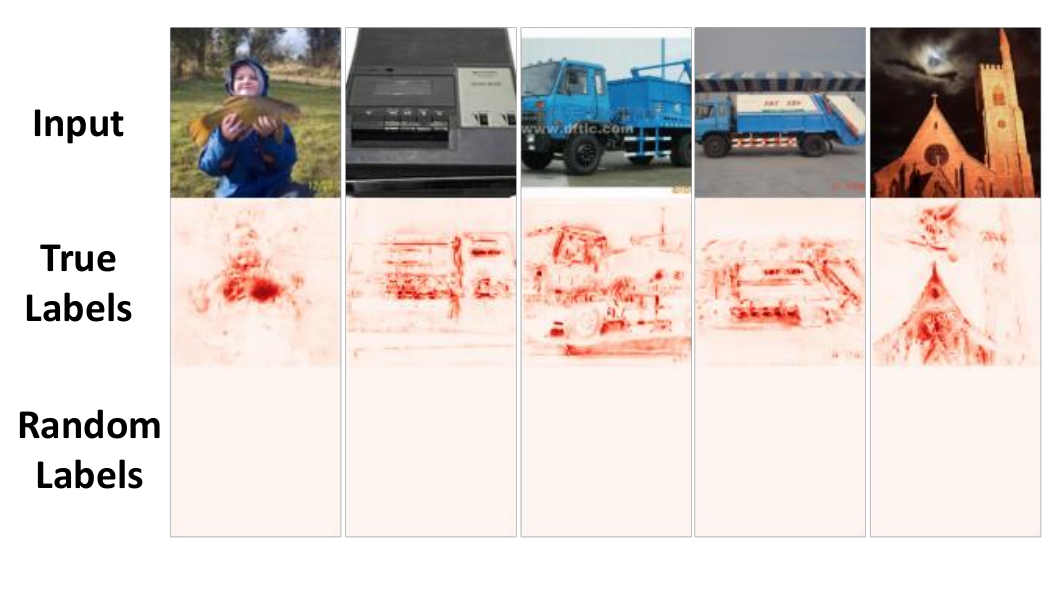}
    \caption{Explanation for a true model vs. model trained on random labels. The true model is trained with elastic net regularized adversarial training.}
    \label{sanity_label}
    \vspace{-3mm}
\end{figure}

\subsubsection{Incorporation of stronger adversarial attacks}
\label{sec:mifgsm}
\begin{figure}[tp!]
    \centering
\includegraphics[width=\linewidth]{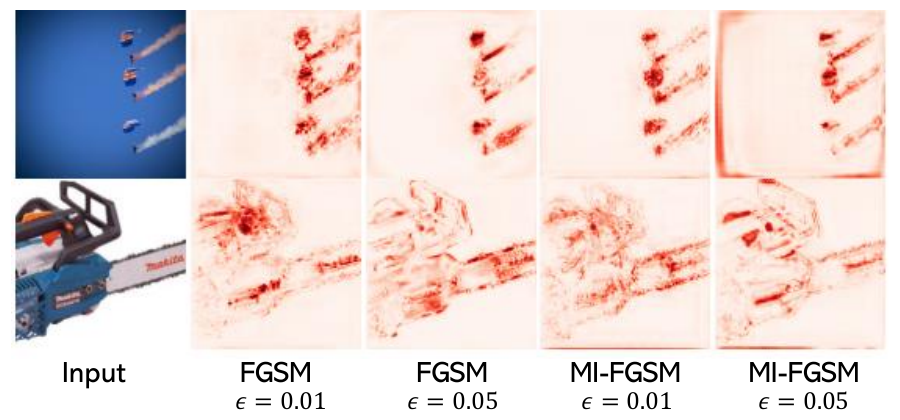}
    \caption{Qualitative comparison between FGSM and MI-FGSM.}
    \label{fig:mifgsm}
\end{figure}

One intuitive idea that can potentially further improve the performance of our method is through the incorporation of stronger adversarial attacks. To see how the saliency maps as we use more powerful attack tools, we compare the $L_1$-regularized adversarial training with Momentum iterative fast gradient sign method (MI-FGSM)~\cite{dong2018boosting} and vanilla FGSM. MI-FGSM leverages momentum-based iterative algorithms to better optimize the perturbation, which is a stronger adversarial attack technique. The results are shown in Fig.~\ref{fig:mifgsm}. Here we showcase two examples with different magnitudes of the attacks. Both FGSM and MI-FGSM generate saliency maps with sparseness and of pretty good visual quality. Therefore, we conclude that momentum-based adversarial training could improve the adversarial robustness of the network while performing similarly on the sparsity of saliency maps.
\subsubsection{Comparison with post-processing methods} 
\label{sec:fidelity}
We further illustrate the results of our proposed methods compared to the post-processing sparsification strategy by measuring the fidelity of each method. We employed $\text{AOPC}_\text{MoFR}$~\cite{samek2016evaluating} at different steps as the fidelity measure and examined the fidelity of sparsified simple gradient, Sparsified-Smoothgard, and Sparse Moreaugrad~\cite{zhang2023moreaugrad}. As the absolute value of fidelity can be affected by the confidence of the network output, it can be unfair to compare the absolute value for different networks. We reported the relative drop in fidelity compared with a based method. Meanwhile, to eliminate the influence of smoothing operation~\cite{yeh2019fidelity} and to only focus on sparsification, we compared sparsified simple gradient with simple gradient and compared Sparsified-SmoothGard and Sparse Moreaugrad with SmoothGrad. The results are summarized in Table~\ref{tab:fid}. These results suggest that post-processing methods often enforce higher sparsity at the expense of lower fidelity. On the other hand, our proposed in-processing scheme preserves the fidelity of the interpretation maps as it outputs the simple gradient maps, which are the most fundamental saliency maps. 

\begin{table}[tp]
\centering
\renewcommand\arraystretch{1}
    \caption{Fidelity of post-processing methods. The reported score is the relative drop compared with base interpretation methods.}
\begin{center}
\vspace{-0.5cm}
\label{tab:fid}
\footnotesize
\begin{tabular}{c|ccc}
\toprule
Metric & \multicolumn{3} {c} {$\Delta \text{AOPC}_\text{MoFR}$ (\%) ($\downarrow$)} \\
\midrule
Steps & 20 & 50 & 100  \\
\midrule
Sparsification&2.75&18.93&36.16\\
Sparsified-SmoothGrad&25.34&24.84&18.13\\
Sparse MoreauGrad&11.29&32.32&45.61\\
\bottomrule
\end{tabular}
\end{center}
\vspace{-0.5cm}
\end{table}

\subsubsection{Integrating with post-processing methods}
\label{sec:img_3}
\begin{figure*}[tp!]
    \centering
    \includegraphics[width=1\linewidth]{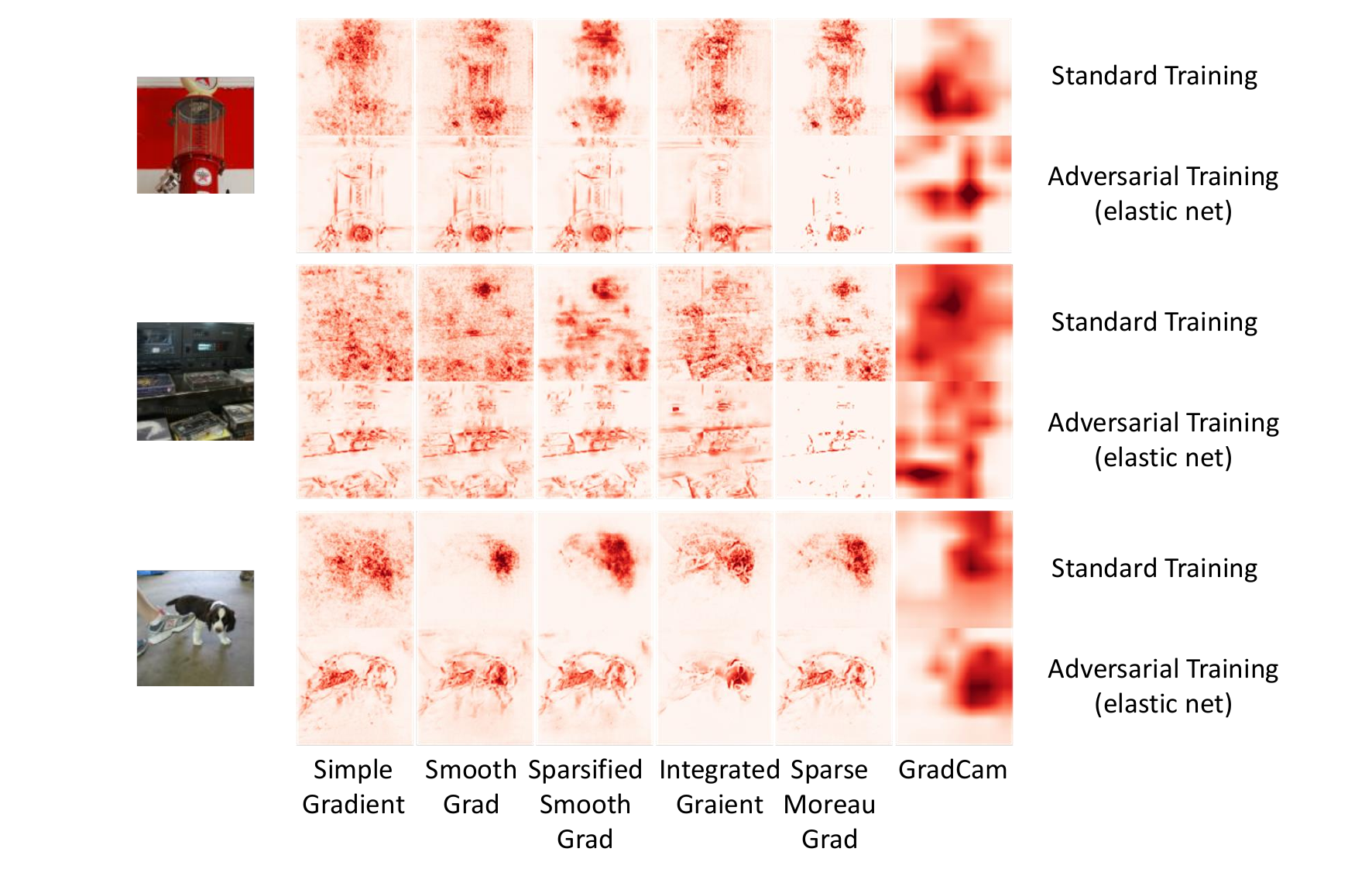}
    \caption{Integrating adversarial training and post-processing methods for generating saliency maps.}
    \label{fig:post_supp}
\end{figure*}

Our adversarial training methodology is an in-processing scheme, which can be more faithful to the original simple gradient map compared with the post-processing scheme. Moreover, if we only care about the visual quality of the saliency maps, these two techniques are orthogonal and may have an additive effect. To illustrate this, we presented saliency maps of other interpretation methods in addition to simple gradient, using the network trained with adversarial training. The results are shown in Fig.~\ref{fig:post_supp}. Although post-processing methods such as smoothing and sparsification can significantly improve the visual quality of saliency maps for standard training, it seems to have minor improvements on Adversarial training, as the original saliency map is already clean and sparse. Nevertheless, performing adversarial training and post-processing together still produce the most visually coherent map.

\subsubsection{Sanity Check}
\label{sec:img_4}
Following~\cite{adebayo2018sanity}, we conducted sanity checks to further show our in-processing scheme could maintain the fidelity of the interpretation maps. We performed both the model parameter randomization test and the data randomization test.

\begin{figure}[tp!]
    \centering
    \includegraphics[width=1\linewidth]{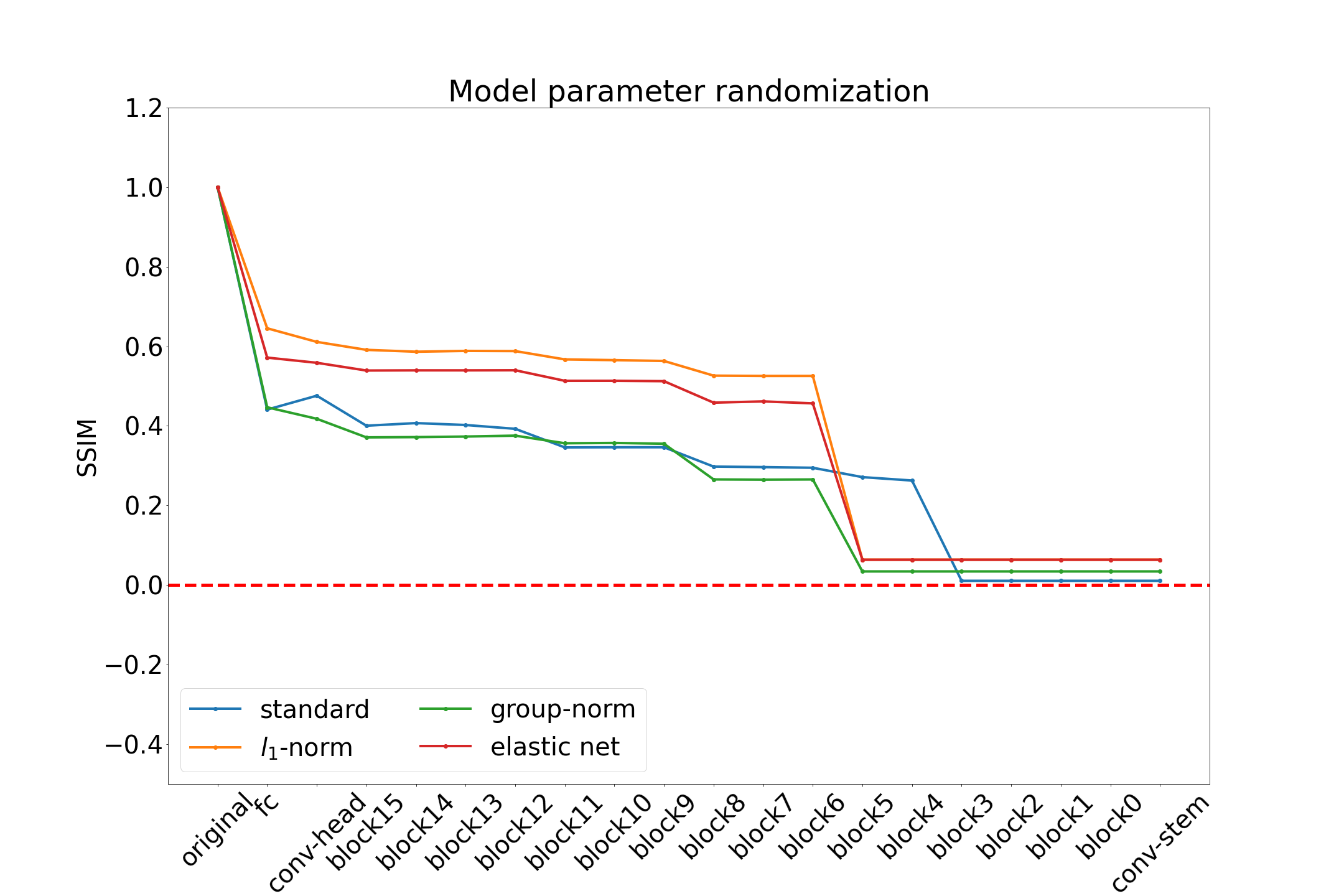}
    \caption{SSIM between saliency maps before and after cascading randomization on the ImageNette dataset.}
    \label{sanity_ssim}
    \vspace{-3mm}
\end{figure}

\begin{figure*}[tp!]
    \centering
    \includegraphics[width=1\linewidth]{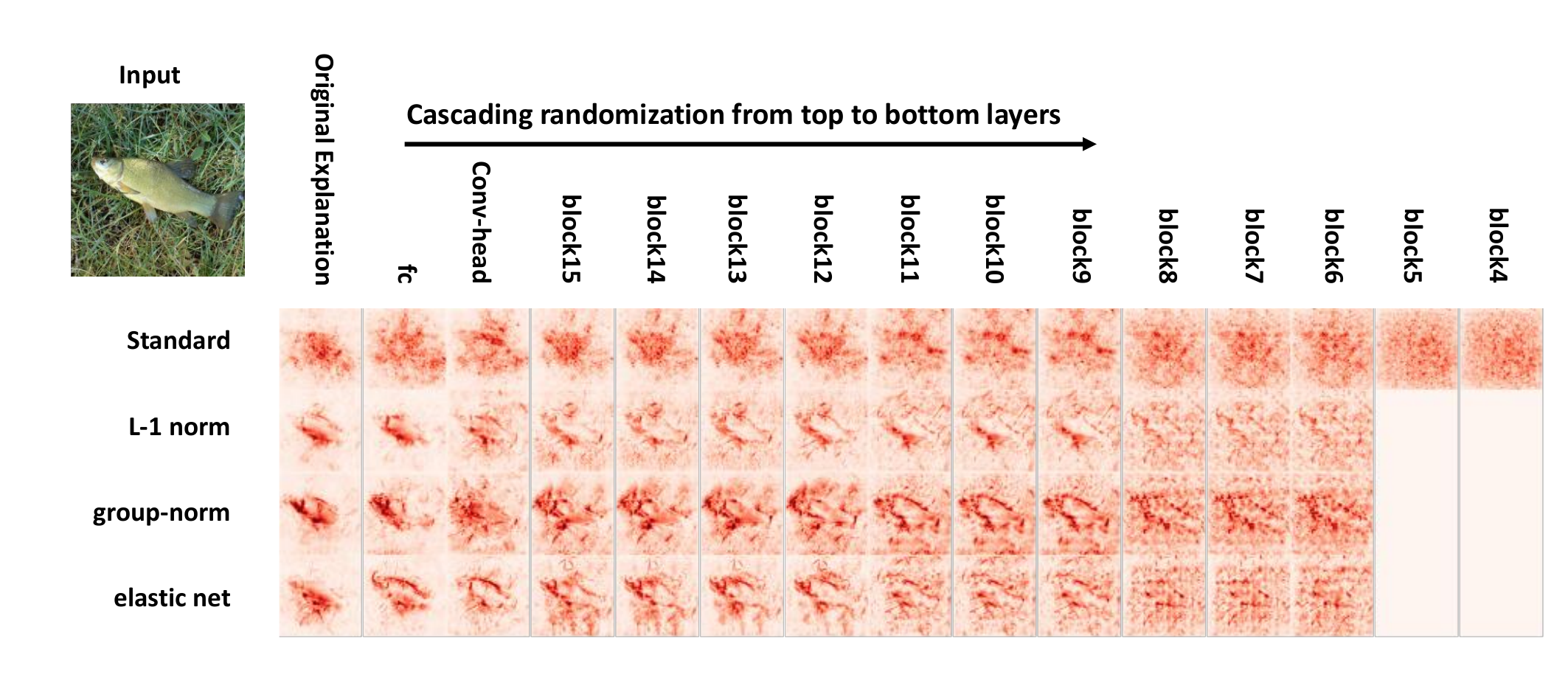}
    \caption{Cascading randomization on EfficientNet-B0 (ImageNette). The first column corresponds to the original explanations for the tench. Progression from left to right indicates complete randomization of network weights. Some saliency maps look blank because the importance scores for all pixels are uniformly small.}
    \label{fig:vis_cifar}
\end{figure*}

\noindent \textbf{Model parameter randomization test.} For the model parameter randomization test, we conducted cascading randomization on the ImageNette data. We computed the SSIM scores comparing the saliency maps after the progressive randomization of layers from the output layer toward the input layer with the original saliency map (Fig.~\ref{sanity_ssim}). The decay trend shows the trained weight is important for the saliency map patterns. We also visualized the saliency maps of cascading randomization (Fig.~\ref{fig:vis_cifar}). In all the cases, the saliency maps of the networks with fully reinitialized weights looked blank, implying no specific parts on the input image play major roles in the decision of the network.

\noindent \textbf{Data randomization test.} In the data randomization test, we randomly permuted all labels and retrained a network with the noisy dataset. The numerical results are shown in Fig.~\ref{sanity_label}. We found that when using adversarial training to train a model with random labels, the model outputs exactly the same prediction disregarding the input images. As a result, the simple gradient saliency maps are composed of all zero values. This shows the interpretation methods indeed try to catch the relationship between inputs and outputs, which is essential in the application of interpretation methods to high-stake phenomena understanding. 

\subsubsection{Visualization of interpretation attacks}
\label{sec:img_5}
\begin{figure*}[tp!]
    \centering
    \includegraphics[width=1\linewidth]{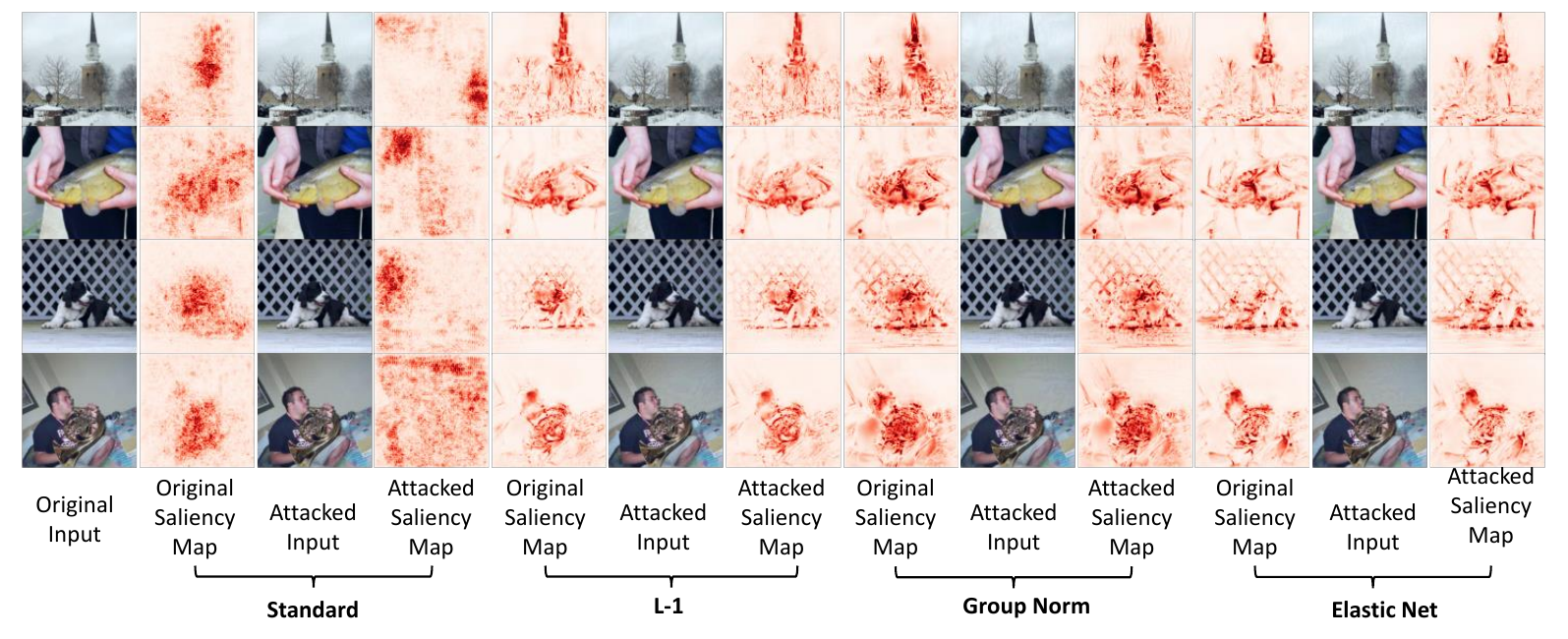}
    \caption{Visualization of attacked image and attacked saliency maps.}
    \label{fig:attack}
\end{figure*}

We further illustrated the robustness of interpretation maps induced by adversarial training. To do so, we visualized several examples showing the interpretation maps before and after the attack. The results are shown in Fig.~\ref{fig:attack}. The result shows the simple gradient saliency map can be very vulnerable to small adversarial perturbation. Although the attacked images have no perceptible difference from the original one, the saliency map can be pretty different. On the other hand, adversarial training can significantly improve the robustness of the interpretation method.

\subsection{Harmonization with gaze maps}
\subsubsection{Comparison with vanilla $L_2$ regularization}
\label{sec:img_6}
\begin{figure}[tp!]
    \centering
\includegraphics[width=1\linewidth]{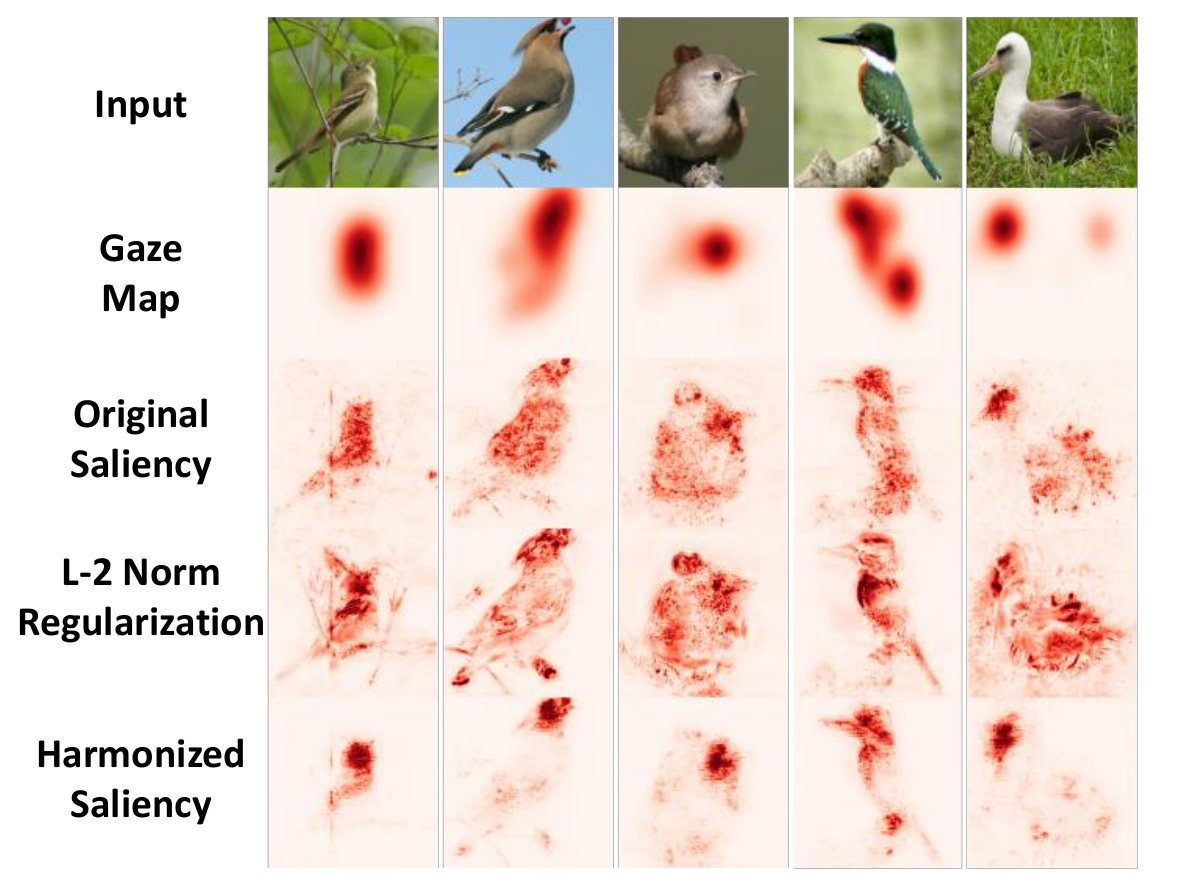}
    \caption{Qualitative comparison between $L_2$-norm regularized adversarial training and our harmonization strategy.}
    \label{fig:cub_l2}
\vspace{-0.5cm}
\end{figure}

Our interpretation harmonization applies a weighted $L_2$-norm to regularize the adversarial training. One potential baseline is to use the vanilla $L_2$-norm to penalize the training so that only important features will be highlighted. Therefore, we reported the top-$k$ overlap of saliency maps generated with $L_2$-norm regularized adversarial training and the gaze map (Table~\ref{tab:harmonize_l2}). We also visualized the saliency maps in Fig.~\ref{fig:cub_l2}. We discovered that vanilla $L_2$-norm regularization cannot help align saliency maps with human attention. As for complex images, neural networks make decisions in a way different from human beings. The features valued by the network can be very different from domain experts, although these features may also be helpful. Our interpretation harmonization strategy can narrow down the gap, making the neural network ``think'' like a human.

 \begin{table}[tp]
\centering
\renewcommand\arraystretch{1}
    \caption{Top-k overlap (\%) between gaze map and saliency map with $L_2$-norm-regularized adversarial training on CUB-GHA dataset. }
\begin{center}
\vspace{-0.5cm}
\footnotesize
\label{tab:harmonize_l2}
\begin{tabular}{c|ccccc}
\toprule
$\epsilon$ & 0 & 0.1 & 0.5 & 1.0 & 5.0 \\
\midrule
top 5\% overlap&33.42&32.92&33.52&34.22&31.83\\
top 10\% overlap&41.81&41.11&40.89&42.05&39.84\\
\bottomrule
\end{tabular}
\end{center}
\end{table}


\end{document}